\documentclass[10pt,twocolumn,letterpaper]{article}

\usepackage{iccv}
\usepackage{times}
\usepackage{epsfig}
\usepackage{graphicx}
\usepackage{amsmath}
\usepackage{amssymb}
\usepackage{color,soul}
\usepackage{xcolor}
\usepackage{xfrac} 
\usepackage{booktabs}
\usepackage{afterpage}
\usepackage{placeins}

\usepackage[pagebackref=true,breaklinks=true,letterpaper=true,colorlinks,bookmarks=false]{hyperref}

\usepackage{cleveref}

\iccvfinalcopy 

\def\iccvPaperID{****} 

\ificcvfinal\fi

\begin{document}

\title{Contrasting Contrastive Self-Supervised Representation Learning Pipelines}

\author{Klemen Kotar$^1$, Gabriel Ilharco$^2$, Ludwig Schmidt$^2$, Kiana Ehsani$^1$, Roozbeh Mottaghi$^{1,2}$ \\
$^1$PRIOR @ Allen Institute for AI, $^2$ University of Washington

}

\newif\ifcomments
\commentstrue 
\ifcomments
    \providecommand{\gamaga}[1]{{\protect\color{teal}{\bf [GI: #1]}}}
    \providecommand{\kiana}[1]{{\protect\color{purple}{\bf [kiana: #1]}}}
    \providecommand{\klemen}[1]{{\protect\color{blue}{\bf [klemen: #1]}}}
   \providecommand{\todo}[1]{{\protect\color{red}{\bf [ToDo: #1]}}}
\else
    \providecommand{\gamaga}[1]{}
    \providecommand{\klemen}[1]{}
    \providecommand{\kiana}[1]{}
    \providecommand{\todo}[1]{}
\fi

\definecolor{cl1}{rgb}{0.235,0.47,0.847}
\definecolor{cl2}{rgb}{0.415,0.658,0.309}
\definecolor{cl3}{rgb}{0.8,0,0}
\definecolor{cl4}{rgb}{0.945,0.76,0.196}

\maketitle
\ificcvfinal\thispagestyle{empty}\fi

\begin{abstract}
In the past few years, we have witnessed remarkable breakthroughs in self-supervised representation learning. Despite the success and adoption of representations learned through this paradigm, much is yet to be understood about how different training methods and datasets influence performance on downstream tasks. In this paper, we analyze contrastive approaches as one of the most successful and popular variants of self-supervised representation learning.  We perform this analysis from the perspective of the training algorithms, pre-training datasets and end tasks. We examine over 700 training experiments including 30 encoders, 4 pre-training datasets and 20 diverse downstream tasks.
Our experiments address various questions regarding the performance of self-supervised models compared to their supervised counterparts, current benchmarks used for evaluation, and the effect of the pre-training data on end task performance. Our Visual Representation Benchmark (ViRB) is available at: \url{https://github.com/allenai/virb}.
\end{abstract}
\section{Introduction}
Learning compact and general representations that can be used in a wide range of downstream tasks is one of the holy grails of computer vision. In the past decade, we have witnessed remarkable progress in learning representations from massive amounts of \textit{labeled} data \cite{alexnet,vgg,resnet}. More recently, \textit{self-supervised} representation learning methods that do not rely on any explicit external annotation have also achieved impressive performance \cite{mocov1,pirl,simclr,byol,swav}. 
Among the most successful approaches are \textit{contrastive} self-supervised learning methods that achieve results close to their supervised counterparts. These methods
typically learn by contrasting latent representations of different augmentations, transformations or cluster assignments of images. With a sufficient amount of transformations and images to contrast, the model is driven to learn powerful representations.

\begin{figure}[tp]
    \centering
    \includegraphics[width=19pc]{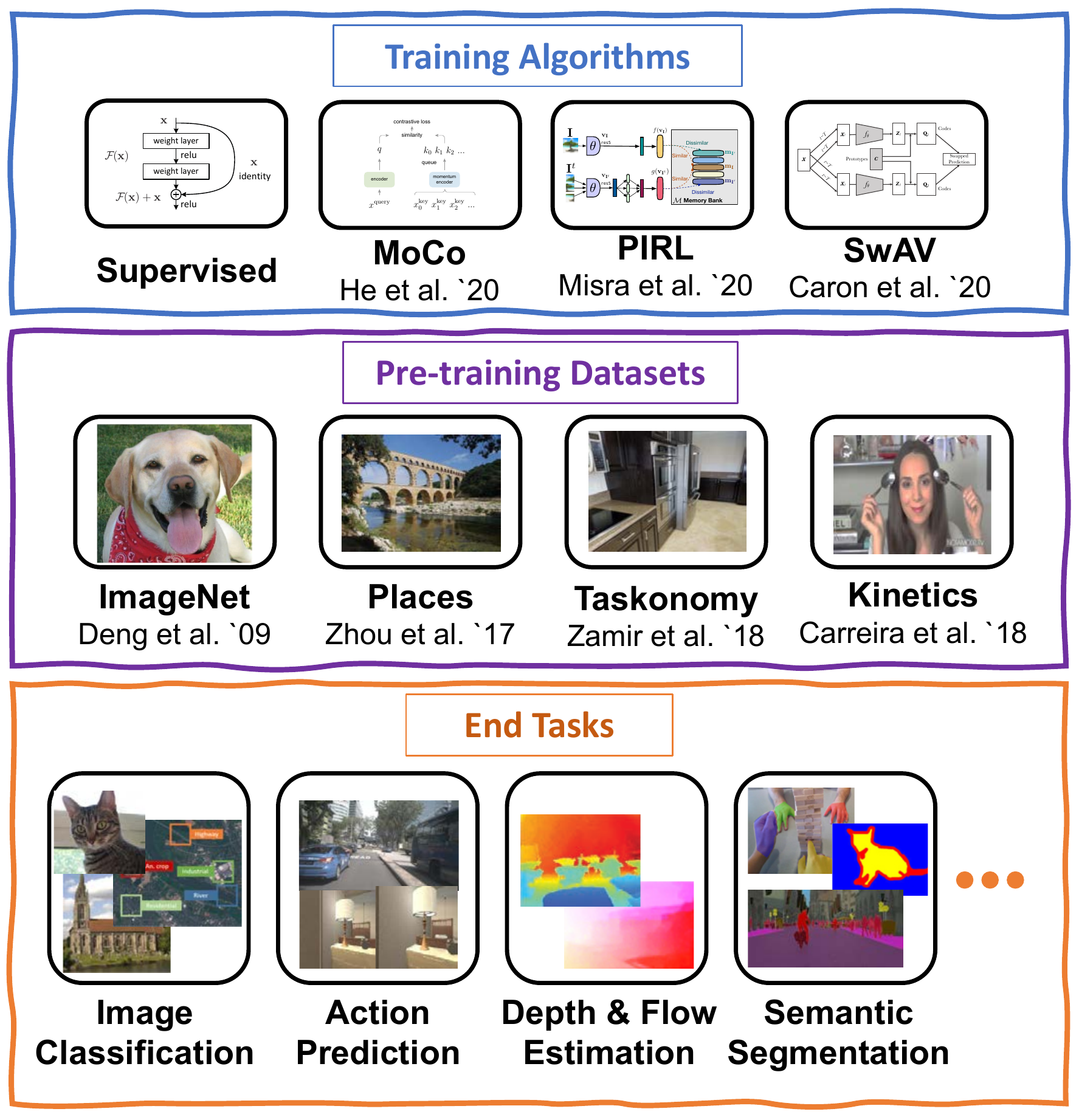}
    \caption{Our goal is to study recently proposed contrastive self-supervised representation learning methods. We examine three main variables in these pipelines: training algorithms, pre-training datasets and end tasks. We consider 4 training algorithms, 4 pre-training datasets and 20 diverse end tasks for this study. }
    \vspace{-0.3cm}
    \label{fig:teaser}
\end{figure} 

The most common protocol for comparing representations learned by self-supervised methods is to pre-train models on a large dataset such as ImageNet \cite{imagenet} without using class labels and then use the learned representations for training end tasks such as image classification, object detection or segmentation. Although this protocol has been widely adopted, it provides an incomplete picture of progress, since the noticeable similarities between common pre-training and end tasks might lead to biased and optimistic estimates of performance. 

In this work, we provide a comprehensive study of representations learned by contrastive self-supervised methods. We explore various alternatives for algorithms, pre-training datasets and end tasks (Figure~\ref{fig:teaser}), covering a total of 735 experiments, using 4 algorithms, 4 pre-training datasets and 20 diverse end tasks. Our goal is to provide answers to the following open questions: (1)  Is supervised learning on ImageNet a good default encoder choice for end tasks? (2) Is ImageNet accuracy a good metric for measuring progress in self-supervised representation learning? (3) How do different training algorithms compare for different end tasks? (4) Does self-supervision provide better encoders for certain types of end tasks? (5) Does the distribution of pre-training data affect the end-task performance? (6) Do we learn poor representations when using highly unbalanced datasets?

We perform an extensive set of experiments to systematically analyze contrastive self-supervision and provide answers to the above questions. We observe a mixture of unintuitive and intuitive results, which better demonstrate the characteristics of contrastive self-supervised models.

\section{Related Work}

\noindent\textbf{Self-supervised representation learning.} To circumvent the need for explicit supervision, various self-supervised approaches have been proposed in previous works. A number of different ``pretext" tasks have been proposed with the goal of training visual encoders, for instance: predicting the spatial configuration of images \cite{doersch2015unsupervised}, colorizing grayscale images \cite{zhang2016colorful}, finding the correct ordering of jigsaw puzzles \cite{noroozi2016unsupervised}, backprojecting to the latent space of GANs \cite{Donahue2016AdversarialFL}, counting primitives \cite{Noroozi2017RepresentationLB}, cross-channel image prediction \cite{Zhang2017SplitBrainAU}, generating image regions conditioned on their surroundings \cite{pathakCVPR16context} and predicting the orientation of an image \cite{gidaris2018unsupervised}. Previous work also explored learning from videos by using ego-motion as supervisory signal \cite{agrawal2015learning,jayaraman2015learning}, tracking similar patches \cite{Wang_UnsupICCV2015}, predicting future frames \cite{Vondrick2015AnticipatingVR} and segmentation based on motion cues \cite{pathak2017learning}. The recent contrastive methods, which are the focus of this study, outperform these approaches and are described next. 

\noindent\textbf{Contrastive representation learning.} Here, we discuss a selection of related contrastive learning methods. Contrastive Predictive Coding (CPC) \cite{CPC} learns a representation by predicting future latent representations using an autoregressive model and a contrastive loss, DIM \cite{DIM} maximizes the mutual information between a region of the input to the encoder and its output, MoCo \cite{mocov1,mocov2} maintains a large memory bank of samples for computing the contrastive loss, SimCLR \cite{simclr,simclrv2} does not use a memory bank and introduces a non-linear transformation between the representation and the loss function, PIRL \cite{pirl} learns similar representations for different transformations of an image, and SwAV \cite{swav} avoids explicit pairwise feature comparisons, contrasting between multiple image views by comparing their cluster assignments. In this paper, we use a subset of the most recent methods that provide state-of-the-art results and have public implementations available. 

\noindent\textbf{Representation learning analysis.} There have been various studies analyzing representations learned via supervised or self-supervised learning. \cite{damour20} analyze the mismatch between training and deployment domains, \cite{taori2020measuring} analyze the robustness to natural data distribution shifts compared to synthetic distribution shifts, \cite{pmlr-v97-recht19a} analyze the generalization capabilities of models trained on ImageNet. \cite{taskonomy} explore the relationships between visual tasks. In contrast to these approaches, we study self-supervised approaches. \cite{vtab} provide a standard benchmark for analyzing the learned representations. \cite{asano2020a} study representations learned at different layers of networks by self-supervised techniques. \cite{purushwalkam2020demystifying} study the effect of invariances such as occlusion, viewpoint and category instance invariances on the learned representation. \cite{tian2020} study the effect of training signals (referred to as ``views") on the downstream task in self-supervised contrastive settings. \cite{Goyal2021SelfsupervisedPO} analyze training self-supervised models on uncurated datasets. \cite{Newell2020HowUI} provide insights about the  utility of self-supervised methods when the number of available labels grows and how the utility changes based on the properties of training data. \cite{ericsson2021selfsupervised} show that on various tasks self-supervised representations outperform their supervised counterpart and ImageNet classification accuracy is not highly correlated with the performance on few-shot recognition, object detection and dense prediction. \cite{sariyildiz2020concept} propose a benchmark to evaluate the representation learning models for generalization to unseen concepts. They evaluate contrastive self-supervised methods as well and show supervised models are consistently better. There are a few concurrent works that analyze representation learning as well. \cite{cole2021does} study the effects of data quantity, data quality, and data domain on the learned representations. \cite{reed2021selfsupervised} sequentially pre-train on datasets similar to the end task dataset and show faster convergence and improved accuracy. \cite{vanhorn2021benchmarking} propose two large-scale datasets and show self-supervised approaches are inferior to supervised ones in these domains. In contrast, we analyze self-supervised contrastive approaches from the perspective of \emph{training algorithms}, \emph{pre-training datasets} and \emph{end tasks}. 

\section{Self-supervision Variables}

Given a set of images $\mathcal{X} = \{x_1, \dots, x_N\}$, the goal of a self-supervised learning algorithm $\Psi$ is to learn parameters $\theta$ of a function $f_\theta$ that maps images $x$ to representations in a continuous latent space. In other words, given an architecture $f$, we learn $\theta = \Psi_f(\mathcal{X})$. The learned representations can then be evaluated on various (supervised) end tasks $\mathcal{D} = \{(\bar{x}_1, y_1), \dots, (\bar{x}_M, y_M)\}$ with pairs of inputs and labels. There are various variables involved in this pipeline. We primarily focus on three variables and their relationship: \emph{training algorithms} $\Psi$, \emph{pre-training datasets} $\mathcal{X}$ and \emph{end tasks} $\mathcal{D}$. Below, we describe each of these variables and the choices for our experiments.


\subsection{Training Algorithms}
 The representation learning algorithms we consider are contrastive self-supervised learning approaches that have recently shown substantial improvements over the previous methods. In this study, we investigate the influence of the training algorithms on the learned representations. We use different algorithms: PIRL \cite{pirl}, MoCov1 \cite{mocov1}, MoCov2 \cite{mocov2} and SwAV \cite{swav}. 
 The reason for choosing these specific algorithms is that they achieve state-of-the-art results on standard end tasks, have a  public implementation available, and do not require heavy GPU memory resources, enabling a large-scale analysis. The list of all 30 encoders is in Appendix~\ref{app:encoders}.
 
\subsection{Pre-training Datasets}
\label{sec:dataset}

The de facto standard used for pre-training with contrastive methods is the ImageNet \cite{imagenet} dataset \cite{pirl,mocov1,simclr,swav}. ImageNet is an object-centric dataset with a balanced number of images for each category. Some works \cite{mocov1,swav} have also used less-curated datasets such as Instagram-1B \cite{wslimageseccv2018}. In this paper, we perform a systematic analysis of the datasets in two dimensions. First, we use datasets with different appearance statistics.
We use Places365 \cite{zhou2017places}, Kinetics400 \cite{kinetics} and Taskonomy \cite{taskonomy} in addition to ImageNet for pre-training. Places is a dataset that is scene-centric and includes images of various scene categories (e.g., stadium and cafeteria). Kinetics is an action-centric dataset and involves videos of activities (e.g., brushing hair and dancing). Taskonomy is a dataset of indoor scene images. Examples from each dataset are provided in Figure~\ref{fig:teaser}.

These datasets are larger than ImageNet. To eliminate the effects of training data size, we subsample these datasets to make them the same size as ImageNet (1.3M images). We uniformly sample from each category of the Places dataset. For Kinetics, we sample at a constant frame rate across all videos.
For Taskonomy, we uniformly sample across the different building scenes. Moreover, to explore the effect of using a pre-training dataset with a mixed distribution of appearance, we randomly select a quarter of each of the aforementioned datasets and combine them to form a dataset with non-uniform appearance statistics.
We refer to this dataset as `Combination'.

\begin{figure*}[tp]
    \centering
    \includegraphics[width=38pc]{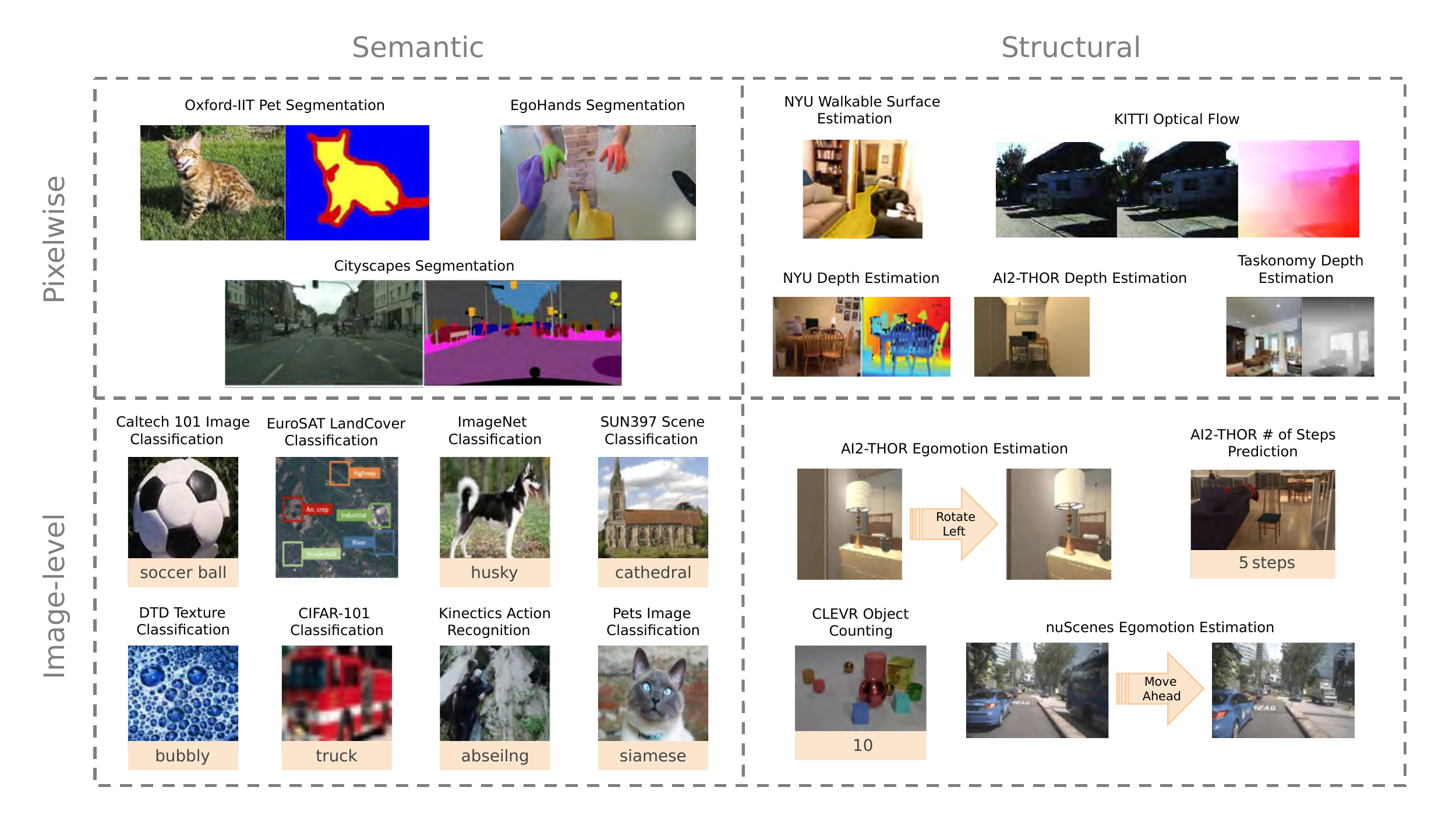}
    \caption{\textbf{End tasks.} We study a diverse set of end tasks. We categorize these tasks according to two characteristics: \emph{semantic} vs. \emph{structural} and \emph{pixelwise} vs. \emph{image-level}. We illustrate an image from each task to show the diversity of visual appearances we consider.}
    \vspace{-0.3cm}
    \label{fig:tasks}
\end{figure*} 

The self-supervised models are typically pre-trained on ImageNet, which is a category-balanced dataset. We also investigate the representations learned on a set of unbalanced datasets. 
We create two unbalanced variations of ImageNet. First, we sample images from each category by linearly increasing the number of samples i.e., we sample one image from category 1, two images from category 2, etc. We refer to this dataset as `ImageNet-\sfrac{1}{2}-Lin' and it consists of $500.5K$ images. In the second variation, the number of samples increases according to an exponential distribution.\footnote{More specifically, we sample $\lambda e^{an+b}$ data points for the $n$-th class, $a$,$b$ and $\lambda$ are chosen so that a single image is sampled from the first class and 1000 images are sampled from the last.} We refer to this unbalanced variation as `ImageNet-\sfrac{1}{4}-Log' and it consists of $250K$ images. To have comparable size datasets, we create smaller balanced variations of the ImageNet dataset by uniformly sampling a quarter and half of the images in each category. We refer to these as `ImageNet-\sfrac{1}{4}' and `ImageNet-\sfrac{1}{2}'. 

\subsection{End Tasks}
\label{sec:end_tasks}

Representations learned from self-supervised methods can be used for various end tasks, such as image classification, object detection and semantic segmentation. Image classification has been considered as the primary end task for benchmarking contrastive self-supervised techniques \cite{Goyal2021SelfsupervisedPO}. Although this task is a reasonable choice for measuring progress, it might not be an ideal representative for various computer vision tasks that are different in nature.
In this study, we consider a wide range of end tasks. To ensure diversity,  we study 20 tasks grouped into four categories based both on the structure of the output and the nature of the task (Figure~\ref{fig:tasks}). The output type of each end task can be classified into two broad categories: \emph{image-level} and \emph{pixelwise}. The former involves reasoning about a region in the image or the entire image, while the latter reasons about each pixel.\footnote{While not the focus of our work, some tasks do not fit into these two categories, e.g. generating future human poses.} Within each category, we consider two categories of tasks based on their nature: \emph{semantic} and \emph{structural}. Semantic tasks are the ones that associate semantic information such as category labels to image regions (e.g., semantic segmentation or image classification). Structural tasks, on the other hand, provide information about some structure in the image (e.g., depth estimation). We note that the boundary between these two types of tasks can become blurry and some tasks can be considered both structural and semantic (e.g., walkable surface estimation). We put these tasks in the closest category. Hence, we have four types of tasks in total:

\begin{itemize}
    \item \textbf{Semantic Image-level.} In these tasks, we provide semantic labels for a region or the entire image. Examples include image classification (e.g., ImageNet classification) and scene classification (SUN397 \cite{sun397} classification). This is the most populated category since most common vision tasks fall into this category.
    \item \textbf{Structural Image-level.} These tasks reason about some structural, global information in images. Example tasks in this category are counting (CLEVR-Count \cite{vtab}) and egomotion estimation (estimating car movements in nuScenes \cite{nuscenes}).
    \item \textbf{Semantic Pixelwise.} In contrast to the two previous categories, the output is pixelwise. The goal is typically to assign a semantic label to each pixel in an image. Semantic segmentation of images in Cityscapes dataset \cite{cityscapes} and hand segmentation in EgoHands \cite{egohands} dataset are example tasks in this category. 
    \item \textbf{Structural Pixelwise.} The fourth category involves providing pixelwise predictions for structural properties in a scene. Examples include estimating pixelwise depth in the AI2-THOR \cite{ai2thor} framework and walkable surface estimation in the NYU Depth V2 \cite{nyudepth} dataset. 
\end{itemize}

Figure~\ref{fig:tasks} illustrates all tasks and their corresponding categories.
More details on the task formulations and their datasets are in Appendix~\ref{app:endtasks}.

\section{Architecture Details}
With the goal of conducting a controlled study, we fix as many variables as possible, and use the standard PyTorch \cite{pyTorch} ResNet50 architecture for every encoder studied. Due to the diverse nature of our tasks and their outputs we have to use several different end task network architectures, but we keep them as small and standard as possible. As a result, we might not achieve state-of-the-art results on every end task. However we ensure that our results are good enough to adequately compare the performance of different learned features. In this section, we describe the architectures used for the backbone encoder and each end task in this study.

\subsection{Encoders}
We remove the final (classification) layer from each trained backbone model and use it as the encoder for all of our end task experiments. Our goal is to investigate the learned representation as opposed to evaluating whether it is an effective initialization. Therefore, we keep the backbone frozen and do not fine-tune the encoders for any task.

\subsection{End Task Networks}
The end task network is the section of the model that converts the embedding produced by the encoder into the desired task output. For each end task we have a train and test set. We train the end task network on the train set using a random initialization and then evaluate it on the test set. We use the same set of hyperparameters for each task in all settings. For further details please see Appendix~\ref{app:networks}.
We have 5 different architectures to suit the wide variety of our end task types.

\noindent\textbf{Single Layer Classifier.} This network contains a single fully connected layer. It takes as input the final ResNet embedding and outputs a vector of size $n$, where $n$ is the number of classes for the task. This network is used for all the image-level classification tasks (e.g., scene classification).

\noindent\textbf{Multi Input Fusion Classifier.} This network contains several ``single linear layer modules", each of which processes one image in a sequence. The outputs of these modules get concatenated and passed through a fusion layer. The network takes as input a series of final ResNet embeddings and outputs a vector of size $n$, where $n$ is the number of classes for the task. This network is used for all the image-level classification tasks that take a sequence of images (e.g., egomotion estimation).

\noindent\textbf{U-Net.} This network is a decoder based on the U-Net \cite{U-Net} architecture---a series of consecutive convolutions followed by upsampling and pixel shuffle \cite{shi2016real} layers. After every upsample, the output of an intermediary representation from the ResNet encoder of matching height and width is added via a residual connection. The final output is a tensor of size $h\times w$, where $h$ and $w$ are the height and width of the input image. This network is used for depth prediction.

\noindent\textbf{Siamese U-Net.} This network is a modification of the U-Net network which can support two images as input.
It takes the final embeddings and intermediary ResNet representations from the two images as input, then fuses them together layer by layer with a point convolution and adds them to the decoder after every convolution via a residual connection. This network is used for flow prediction.

\noindent\textbf{DeepLabv3+.} This network is based on the DeepLabv3+ \cite{DeepLabv3+} architecture. It takes as input the output of the $5$th block of the ResNet and uses dilated convolutions and a pyramidal pooling design to extract information from the representations at different scales. The output is then upsampled and is added to the representation from the $2$nd block of the ResNet to recover image structure information. The final output is of size $n\times h\times w$, where $n$ is the number of output channels, $h$ and $w$ are the height and width of the input image. This network is used for pixelwise semantic classification tasks (e.g., semantic segmentation).

\section{Analysis}
In this section, we pose several questions on the relationships across pre-training algorithms, pre-training datasets and the end tasks. We discuss our experiments' design and analyze the results to provide answers to each of these questions. We perform an extensive analysis of the contrastive self-supervised models and discuss the performance trends in different settings. We also investigate which common intuition used in supervised training transfers over to the self-supervised domain. Unless noted otherwise all training algorithms have been used for the experiments. The implementation and training details are provided in Appendix~\ref{app:train_details}. 

\paragraph{(1) Is supervised learning on ImageNet a good default encoder choice? }

\begin{figure}[tp]
    \centering
    \includegraphics[width=\linewidth]{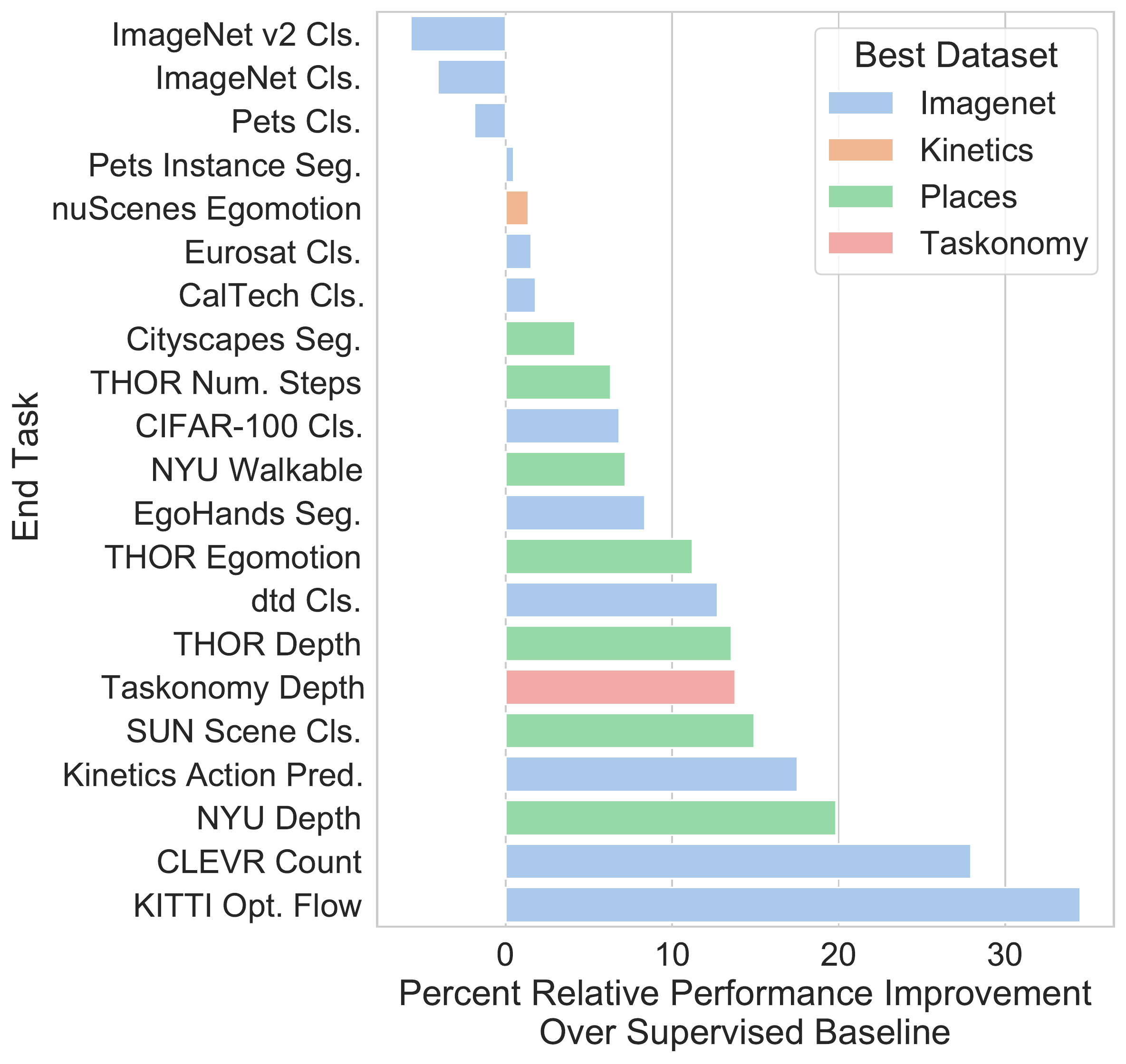}
    \caption{\textbf{Comparison of self-supervised and supervised encoders.} The percentage performance improvement of the self-supervised encoders for each end task is shown. The colors of the bars represent the dataset used for pre-training the best performing self-supervised encoder. The plot shows that the self-supervised encoders are better than an encoder trained on ImageNet in a supervised way except for the three end tasks shown on top, which are ImageNet classification and Pets classification (which is quite similar to ImageNet classification).}
    \vspace{-0.2cm}
    \label{fig:best_encoder}
\end{figure}

A ResNet encoder trained with supervised learning on the ImageNet dataset has become the default backbone for many computer vision models. With the recent rise of self-supervised training algorithms we re-evaluate this assumption. For each of the 20 end tasks, we compare the best performing self-supervised encoder with the encoder trained on ImageNet in a supervised fashion. The performance improvements of self-supervised methods are shown in Figure~\ref{fig:best_encoder}, along with the dataset used for pre-training. For the ImageNet v1 and v2 classification as well as Pets classification (which is very close to the ImageNet task), the supervised model performs the best, but for all other tasks some self-supervised encoder achieves a higher performance. This indicates that a self-supervised model might be a better default option in many scenarios.  

Figure~\ref{fig:best_encoder} also shows that most of the best performing models are pre-trained on ImageNet or Places. Both of these datasets are curated and structured datasets (as opposed to Kinetics and Taskonomy which are unstructured). This might suggest that self-supervised encoders might also benefit more from well-organized training data.


\paragraph{(2) Is ImageNet accuracy a good metric for measuring progress on self-supervised representation learning?}   Most recent works in self-supervised representation learning report the performance of their encoders on different tasks, but the common denominator between them is mostly the ImageNet classification task. We test a variety of encoders on our diverse set of 20 end tasks to observe how well the performance on those tasks correlates with ImageNet classification performance. 

\begin{figure*}[tp]
    \centering
    \includegraphics[width=40pc]{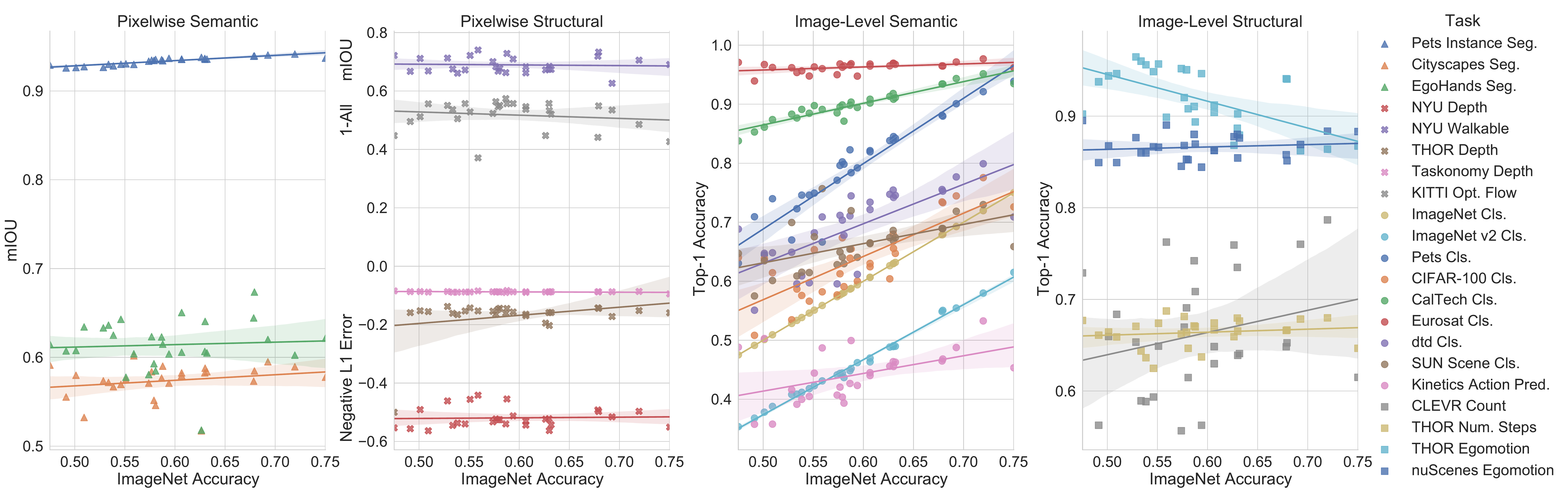}
    \caption{\textbf{Correlation of end task performances with ImageNet classification accuracy.} The plots show the end task performance against the ImageNet top-1 accuracy for all end tasks and encoders. Each point represents a different encoder trained with different algorithms and datasets. This reveals the lack of a strong correlation between the performance on ImageNet classification and tasks from other categories. }
    \vspace{-0.3cm}
    \label{fig:incomp}
\end{figure*} 

Figure~\ref{fig:incomp} contrasts the performance of an encoder on ImageNet versus all other end tasks. The $x$ axis denotes the performance of the learned representation on ImageNet classification and the $y$ axis denotes the performance of the end tasks using the self-supervised encoder. Each point in the plot represents a different encoder obtained by different training algorithms, datasets, etc. 

While we generally observe a strong correlation between the performance on ImageNet classification and other tasks in the same category (semantic image-level), there is a weaker (and sometimes even negative) correlation with tasks in other categories -- refer to Appendix~\ref{app:correlation} for Spearman and Pearson correlation analysis. This indicates that the representations that are suitable for ImageNet classification do not always transfer well to other computer vision tasks. The results for semantic image-level tasks are in line with the findings of \cite{Kornblith2019DoBI}. However, we observe a different trend for the other task types. Note that for some end tasks the performance ceiling might have been reached. Hence, we might not observe a significant difference between different encoders for them.

The fact that we find several tasks that appear to be negatively correlated with ImageNet performance suggests that the encoders that perform quite well on ImageNet might be overfitting to a particular task type and output modality. Interestingly, the category that is most negatively correlated with ImageNet performance is image-level structural tasks, which shares relatively similar network architecture and loss function with ImageNet classification. This provides more evidence that the architecture and the loss function are not the variables that determine the correlations. 

Considering these analyses, ImageNet classification does not appear to be a strong indicator of self-supervised encoder performance for various computer vision tasks.

\paragraph{(3) How do different pre-training algorithms compare for different end tasks?} Two recent strong self-supervised algorithms are MoCov2 \cite{mocov2} and SwAV \cite{swav}. We train several encoders using both algorithms to determine if the trends we observe extend beyond a single algorithm. In addition, this allows us to contrast the MoCov2 and SwAV algorithms to determine if either one is a better fit for certain end tasks.

\begin{figure}[tp]
    \centering
    \includegraphics[width=\linewidth]{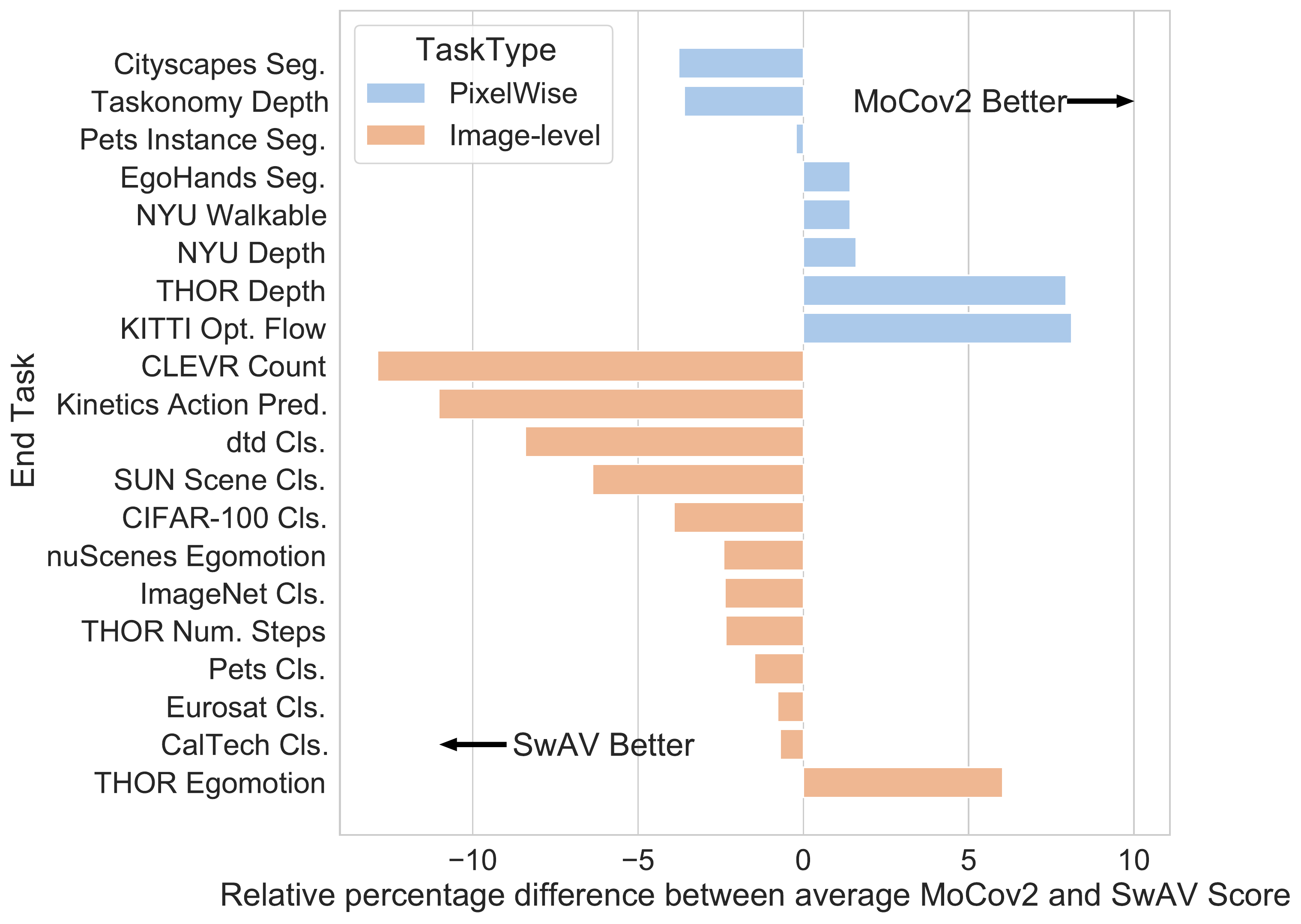}
    \caption{\textbf{Training algorithms and tasks.} For each end task, the difference between the average score of all encoders trained with MoCov2 and the average score of all encoders trained with SwAV is shown. Therefore a negative score indicates that SwAV outperforms MoCov2 on average for a given task and a positive score means the opposite. The scores are unscaled evaluation metrics (accuracy, mIOU or negative L1 error depending on the task). With some exceptions, the plot shows SwAV is generally better at \emph{image-level} tasks, while MoCov2 is better at \emph{pixelwise} tasks.}
    \label{fig:diff}
\end{figure}

For answering this question, we consider encoders trained for 200 epochs on our pre-training datasets. Therefore, we train 10 encoders in total, using our five datasets (ImageNet, Places, Kinetics, Taskonomy, and Combination) by SwAV and MoCov2 methods.
In Figure~\ref{fig:diff}, for each end task, we plot the percentage difference between the average performances of MoCov2 and SwAV encoders. MoCov2 encoders tend to do better at tasks where the output is pixelwise (a notable exception is Cityscapes Segmentation). SwAV models are better at classification tasks, especially semantic classification tasks (here the notable exception is THOR egomotion estimation which is also inversely correlated with ImageNet classification). 
 
Under typical evaluation procedures, SwAV might be considered an absolute improvement over MoCov2, since SwAV outperforms MoCov2 on ImageNet classification. However, our results suggest that this is not a universal fact. This underscores the importance of reporting performance on a diverse and standardized battery of end tasks to show a more comprehensive overview of a model's performance.
 
To investigate if there is some fundamental difference in the representations produced by different encoders, which explains this trend, we compute the \textit{linear Centered Kernel Alignment} (CKA) \cite{CKA} between the outputs of each ResNet block of the MoCov2 and SwAV models. We use a 10,000 image, balanced subset of ImageNet at half resolution for this evaluation. See Appendix~\ref{app:cka} for details. We observe a stronger agreement between the representations in the earlier blocks and later blocks with MoCov2 models, than we do with SwAV models. These trends may suggest that MoCov2 representations are better at capturing low-level information from an image, while SwAV representations are better at capturing higher-level semantic information.

\paragraph{(4) Does self-supervision work better on certain end tasks?}

\begin{figure}[tp]
    \centering
    \includegraphics[width=\linewidth]{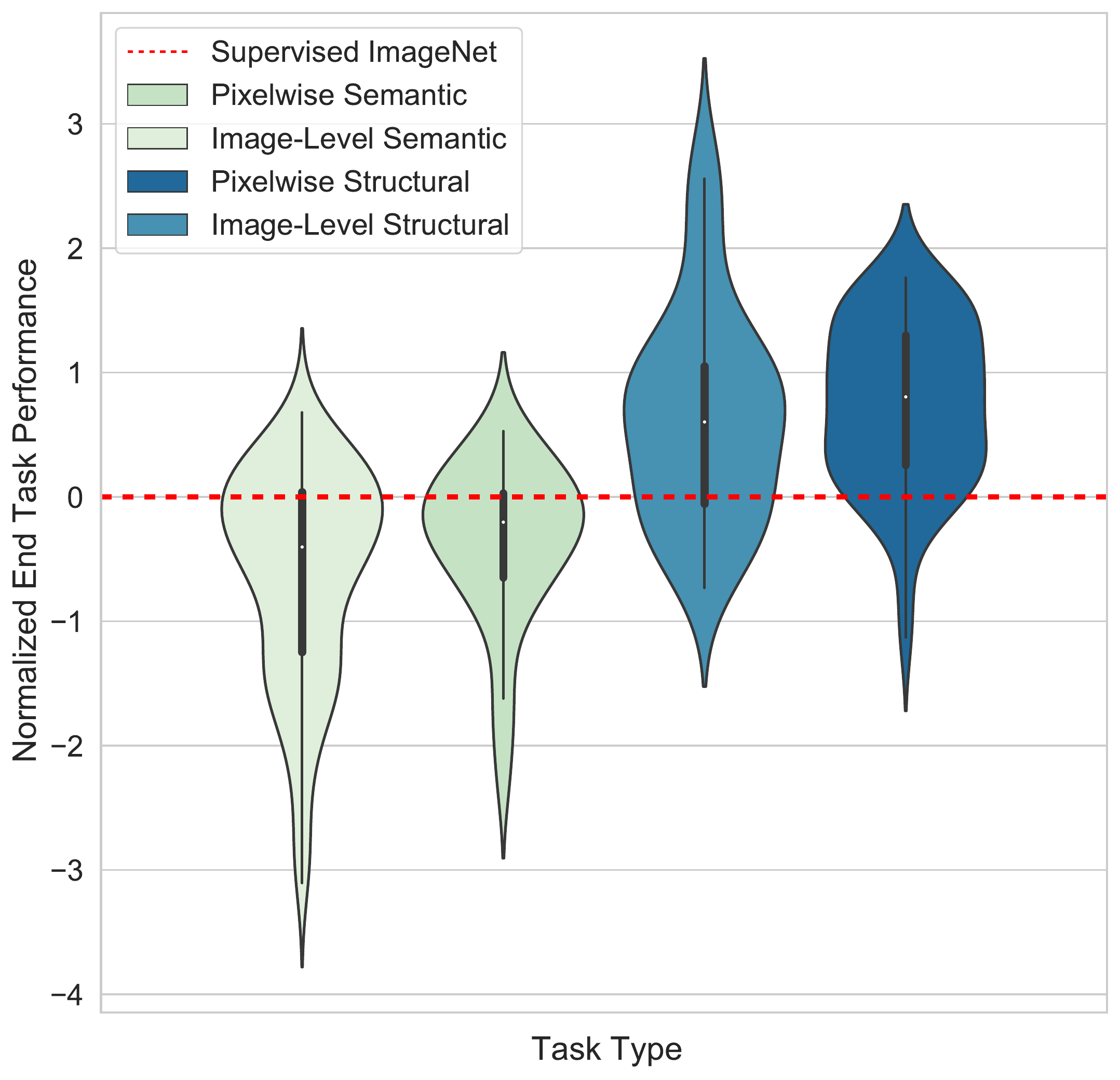}
    \caption{\textbf{Distribution of normalized performances for each category of end tasks.} The performances are normalized by first subtracting the performance of the supervised ImageNet encoder and then dividing by the std. deviation of all the performances for the task. Positive values show superior performance to the supervised ImageNet, and the negative values show otherwise. A larger width means more performance values fall in that range. The plot shows \emph{structural} tasks benefit more from self-supervision.}
    \vspace{-0.2cm}
    \label{fig:violin}
\end{figure}

Pre-trained encoders are used for a variety of applications in computer vision, yet most reported results focus on improvements obtained on semantic tasks such as image classification, object detection and instance segmentation \cite{swav,Goyal2021SelfsupervisedPO}. We would like to obtain a general picture of how well self-supervised encoders perform across each individual task category. Since end tasks use different success metrics, we use a normalization scheme to effectively compare them. In Figure~\ref{fig:violin} we take every performance metric obtained by a self-supervised encoder on an end task and subtract the score obtained by the supervised representation trained on ImageNet. Note that this indicates that the points with positive values outperform the supervised baseline. We then further normalize these values by dividing them by their standard deviation.

Figure~\ref{fig:violin} indicates that \emph{structural} tasks receive a greater benefit from using a self-supervised encoder. Note that the relatively large standard deviation in this plot is due to including self-supervised encoders trained on datasets and algorithms that might not be the best match for the given task type.
Note that this plot does not conflict with our observation in Figure~\ref{fig:best_encoder} on the good performance of self-supervised encoders on \emph{semantic} tasks.
As shown in Figure~\ref{fig:best_encoder}, a self-supervised model outperforms the supervised baseline on all but three semantic image-level tasks.

\paragraph{(5) Does training with images from a similar domain improve performance?}

We hypothesize that using a pre-training dataset similar to the end task's will produce a better encoder.
We choose 4 datasets to test this hypothesis: two structured (ImageNet and Places365) and two unstructured (Taskonomy and Kinetics400). We train two encoders on each of them (MoCov2 and SwAV, the best performing algorithms) and pair each pre-training dataset with an end task using either a dataset in the similar domain as the pre-training data (SUN397 \cite{sun397} classification for Places265 \cite{zhou2017places} and Caltech101 \cite{caltech} classification for ImageNet \cite{imagenet}) or using a subset of the same dataset (action prediction for Kinetics400 and depth estimation for Taskonomy).

\begin{figure}[tp]
    \centering
    \includegraphics[width=0.98\linewidth]{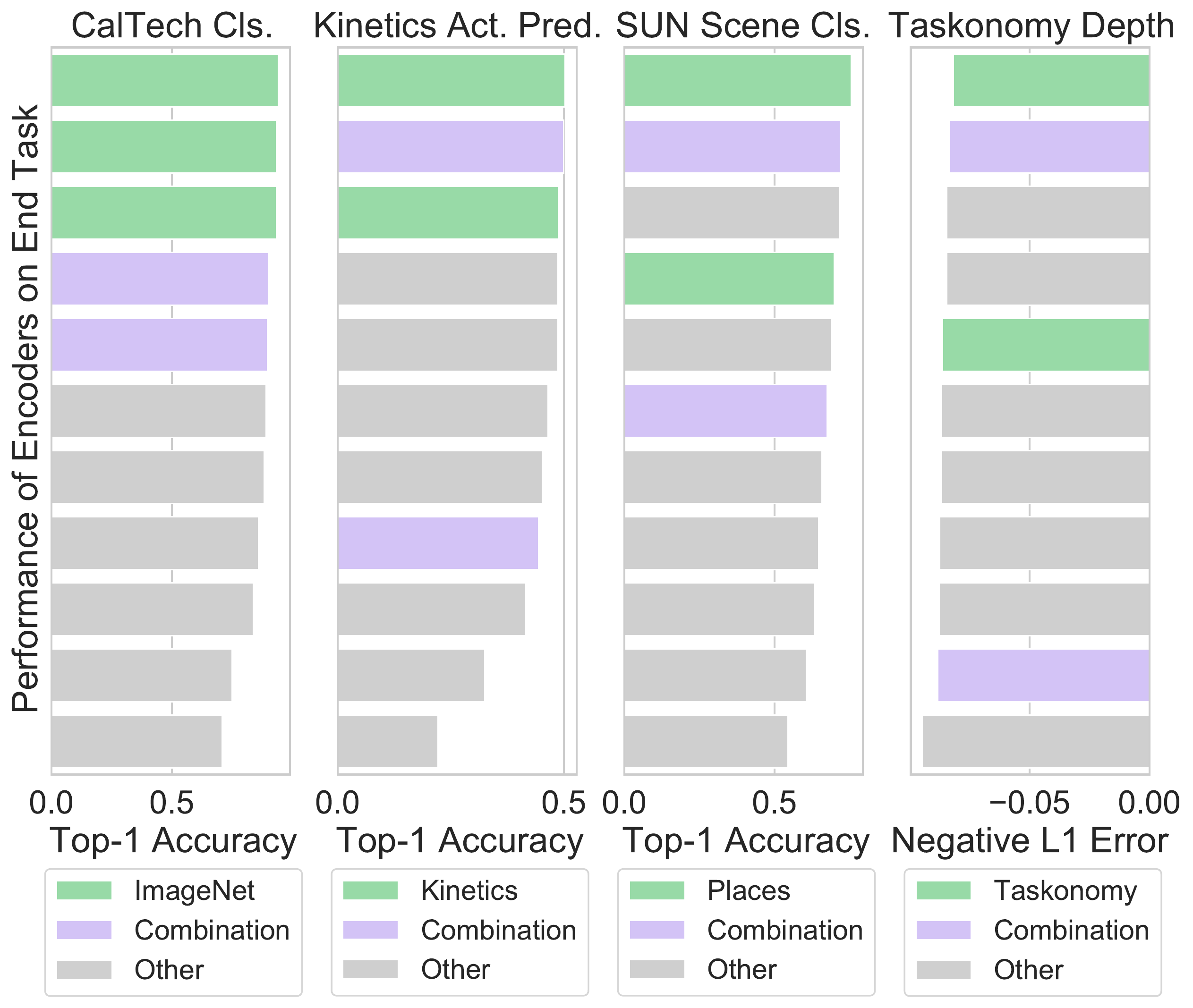}
    \caption{\textbf{Similarity of the pre-training datasets and end tasks.} Performance of all encoders on selected end tasks is shown. Each bar represents a different encoder. The green bars represent encoders pre-trained on a dataset similar to/same as the end task dataset. The purple bars represent the encoders pre-trained on `Combination'. Encoders pre-trained on similar/same datasets have the highest score. Moreover, those encoders are superior to the encoders trained on Combination, which includes not only a subset of that dataset, but also images from other datasets.}
    \vspace{-0.4cm}
    \label{fig:similar}
\end{figure}

In Figure~\ref{fig:similar} we plot the end task performance of MoCov2 and SwAV models trained for 200 epochs on the pre-training datasets mentioned above. The green bars indicate the encoders trained on a dataset that is similar to the end task data, while the gray bars indicate encoders trained on other datasets. The purple bars indicate the encoders trained on the `Combination' dataset (referred to in Section~\ref{sec:dataset}).

We find that for every task, the best performing encoder is the one trained on a dataset that includes similar data. However, as Figure~\ref{fig:similar} shows, the training dataset alone is not enough to determine which encoder will perform the best, as the algorithms also impact the performance. 

We observe that training on `Combination' does not produce a model that excels at every task, therefore, simply combining different datasets with different appearance distributions might not be a good strategy for self-supervised training. Note that the combination dataset still benefits from including images similar to the end task images.

\paragraph{(6) Do we learn poor representations if we use unbalanced ImageNet?}

 Here, we evaluate the learned representations in scenarios where we use unbalanced data for pre-training the encoders. Using unbalanced data better mimics real-world data distributions which are typically long-tailed \cite{openlongtailrecognition}.

We consider two unbalanced subsets of ImageNet (ImageNet-\sfrac{1}{2}-Lin and ImageNet-\sfrac{1}{4}-Log) described in Section~\ref{sec:dataset}, and two correspondingly sized balanced subsets (ImageNet-\sfrac{1}{2} and ImageNet-\sfrac{1}{4}). Encoders are trained on each of the four ImageNet subsets using SwAV and MoCov2 for 200 epochs each, to produce 8 encoders, which are tested on the 20 end tasks. We fit a factorial ANOVA model to the end task results and find no evidence that pre-training on a balanced datasets produces a better encoder. We find that a model being pre-trained on ImageNet-\sfrac{1}{2}-Lin is not a statistically significant predictor of model performance (p-value = 0.0777), while a model being trained on ImageNet-\sfrac{1}{4}-Log is (p-value = 0.0101) with an average end task score improvement of 1.53\%. This presents weak evidence that pre-training on a heavily unbalanced dataset with contrastive learning might even produce an encoder better suited for the end tasks studied in this work. For further details see Appendix~\ref{app:anova}.

\section{Discussion}
\vspace{-0.2cm}
Here we provide a summary of the analysis. First, we showed that a backbone trained in a supervised fashion on ImageNet is not the best encoder for end tasks other than ImageNet classification and Pets classification (which is a similar end task). Second, we showed that in many cases there is little to no correlation between ImageNet accuracy and the performance of end tasks that are not \emph{semantic image-level}. Third, we showed different training algorithms provide better encoders for certain classes of end tasks. More specifically, MoCov2 proved better for \emph{pixelwise} tasks and SwAV showed better performance on \emph{image-level} tasks. Fourth, we showed that \emph{structural} end tasks benefit more from self-supervision compared to \emph{semantic} tasks. Fifth, we showed pre-training the encoder on the same or similar dataset to that of the end task provides higher performance. This is a well-known fact for supervised representation learning, but it was not evident for self-supervised methods that do not use any labels. Sixth, we showed that representations learned on unbalanced ImageNet is as good or even slightly better than representations learned from balanced data. The current study has some shortcomings that are noted below:

\noindent\textbf{Empirical study.} Our conclusions are based on empirical results. This has two major implications. First, there is no theoretical justification for the results. Second, due to computation limits and the wide range of parameters and variables involved in these types of approaches, our study does not cover all aspects related to contrastive self-supervised representation learning. 

\noindent\textbf{Task dichotomy.} The task categorization that we studied is based on the type of output and information they capture. There are several other ways of grouping these tasks that are not studied here and are left for future work. 

\noindent\textbf{Variables.} We focused only on three variables in the representation learning pipeline, namely, training algorithms, pre-training datasets and end tasks. There are various other factors involved in the representation learning pipeline such as network architectures and computational efficiency that are not addressed in this study.

\noindent\textbf{Frozen backbone.}  We did not fine-tune the encoders during training for end tasks. A future direction can be exploring the trends when the encoder is fine-tuned as well. 

\vspace{-0.2cm}
\section{Conclusion}
\vspace{-0.1cm}
We studied contrative representation learning as one of the most successful approaches proposed for self-supervision. Our focus was mainly on three variables in representation learning pipelines, namely, training algorithm, pre-training dataset and end task. Our rigorous analysis resulted in interesting findings about the interplay of these variables. We hope our study provides better insights for future research in this vibrant and impactful domain.

{\small
\noindent\textbf{Acknowledgments:} We would like to thank Luca Weihs for discussions about the statistical analyses.}

{\small
\bibliographystyle{ieee_fullname}
\bibliography{egbib}
}

\clearpage

\appendix
\section*{Appendix}

\section{End Tasks}
\label{app:endtasks}
The descriptions of all end tasks are provided below. \textcolor{cl1}{Semantic Image-level}, \textcolor{cl2}{Structural Image-level}, \textcolor{cl3}{Semantic Pixelwise}, and \textcolor{cl4}{Structural Pixelwise} tasks are shown with different colors.  
\newline

\noindent\textcolor{cl1}{$\bullet$} \textbf{ImageNet Cls.}~\cite{imagenet} - This is a 1000 class natural image classification task. The dataset includes a variety of categories, such as coffee mugs, drum, and fire engine. The images are of varying resolution but they are all resized to 224$\times$224.

\noindent\textcolor{cl1}{$\bullet$} \textbf{ImageNet v2 Cls.}~\cite{pmlr-v97-recht19a} - This is a natural image classification task with the same set of categories as ImageNet. This task has the same train set as ImageNet, but has a re-collected test set with the same distribution and categories.

\noindent\textcolor{cl1}{$\bullet$}~\textbf{Pets Cls.} \cite{parkhi12a} - This is a natural image classification task with images of cats and dogs. There are a total of 37 classes corresponding to breeds of cats and dogs including Persian, Chihuahua and Bengal. The images are of varying resolution but they are all resized to 224$\times$224.

\noindent\textcolor{cl1}{$\bullet$} \textbf{CalTech Cls.}~\cite{caltech} - This is a 101 class natural image classification task with pictures of objects such as planes, chairs, and animals. The images are of varying resolution but they are all resized to 224$\times$224.

\noindent\textcolor{cl1}{$\bullet$} \textbf{CIFAR-100 Cls.}~\cite{Krizhevsky09learningmultiple} - This is a 100 class natural image classification task. The classes include apples, bottles and bicycles. The Images are of size 32$\times$32 but they are all resized to 224$\times$224.

\noindent\textcolor{cl1}{$\bullet$} \textbf{SUN Scene Cls.}~\cite{sun397} - This is a 397 class scenery image classification task. The classes include scene categories such as cathedral, river, or archipelago.

\noindent\textcolor{cl1}{$\bullet$} \textbf{EuroSAT Cls.}~\cite{eurosat} - This is a 20 class dataset of satellite imagery classification. The spatial resolution corresponds to 10 meters per pixel and includes categories of different land use, such as Residential, Industrial and Highway. All images are resized to 224$\times$224.

\noindent\textcolor{cl1}{$\bullet$} \textbf{dtd Cls.}~\cite{dtd} - This is a 47 class dataset of textural imagery classification. Some textures include bubbly, lined and porous. All images are resized to 224$\times$224.

\noindent\textcolor{cl1}{$\bullet$} \textbf{Kinetics Action Pred.}~\cite{kinetics} - This task consists of predicting the action that a person is performing from 6 ordered video frames. The dataset contains 400 classes including bowling and dining. The dataset for this task is 50,000 image frames captured from the Kinetics400 dataset videos at 6 frames per video. All images are resized to 224$\times$224. 

\noindent\textcolor{cl2}{$\bullet$} \textbf{CLEVR Count}~\cite{clevr} - This is a synthetic visual question answering dataset designed to evaluate algorithmic visual reasoning. The task consists of classifying the number of objects in the image.  All images are resized to 224$\times$224.

\noindent\textcolor{cl2}{$\bullet$} \textbf{THOR Num. Steps} - This is a task where the maximum number of forward steps (of 0.25 meters) that a robot in AI2-THOR \cite{ai2thor} can take is predicted from a frame of the robot's point of view. This task is structured as classification, rather than regression, of the images from the simulation and the correct answer will always be between 0 and 4 steps inclusive (thus this task is a 5-way classification). This task is first proposed in this paper. 

\noindent\textcolor{cl2}{$\bullet$} \textbf{nuScenes Egomotion} - This is an egomotion prediction task from two consecutive frames of the nuScenes self driving car dataset \cite{nuscenes}. The types of motion include forward, forward-left and forward-right motion as well as a no motion action. Both frames are resized to 224$\times$224. This task is first proposed in this paper.

\noindent\textcolor{cl2}{$\bullet$} \textbf{THOR Egomotion} - This is an egomotion prediction task from two consecutive frames in the AI2-THOR \cite{ai2thor} simulator. The types of motion include moving  forward, left and right rotation, and looking up and down. Frames are resized to 224$\times$224. This task is first proposed in this paper.

\noindent\textcolor{cl3}{$\bullet$} \textbf{Cityscapes Seg.}~\cite{cityscapes} - This is a semantic segmentation task where every pixel is labeled as one of 20 categories. The images consist of dashboard camera views of cities and roads. The task contains categories such as person, traffic light and sky (there is also a background class for pixels that do not fit into any other category). Crops of size 513$\times$513 sampled from the full image are used during training, and evaluation is done at full resolution.

\noindent\textcolor{cl3}{$\bullet$} \textbf{Pets Instance Seg.} - This is an instance segmentation task on the Pets dataset \cite{parkhi12a}, where each image contains exactly one cat or dog. Each image (and its ground truth instance label) is resized to 224$\times$224.

\noindent\textcolor{cl3}{$\bullet$} \textbf{EgoHands Seg.}~\cite{egohands} - This is an instance segmentation task on a dataset of video frames of human hands performing various tasks. The videos are captured using a Google glass camera and are from the egocentric view of one person performing a task with another person. Each frame has at most 4 hands (the left and right hand of the person wearing the Google glass and the right and left hand of their partner) and each of these has its own associated class (there is also a background class). Crops of size 513$\times$513 sampled from the full image are used during training, and evaluation is done at full resolution.

\noindent\textcolor{cl4}{$\bullet$} \textbf{NYU Depth}~\cite{nyudepth} - This is a pixelwise depth prediction task on a dataset of natural images of building interiors obtained from videos. The images are resized to 224$\times$224 and the output is predicted in meters.

\noindent\textcolor{cl4}{$\bullet$} \textbf{THOR Depth} - This is a pixelwise depth prediction task on a dataset of synthetic images of building interiors produced by the AI2-THOR \cite{ai2thor} simulator. The images are resized to 224$\times$224 and the output is predicted in meters. This task is first proposed in this paper.

\noindent\textcolor{cl4}{$\bullet$} \textbf{Taskonomy Depth}~\cite{taskonomy} - This is a pixelwise depth prediction task on a dataset of natural images of building interiors from a variety of building types. The images are resized to 224$\times$224 and the output is predicted in meters. This is a common task but the dataset split is first proposed in this paper.

\noindent\textcolor{cl4}{$\bullet$} \textbf{NYU Walkable}~\cite{Mottaghi_2016_CVPR} - This is a pixelwise detection task. Each pixel is labeled as walkable (floor, carpet, etc.) or non-walkable (wall, window, ceiling, etc). The dataset consists of images of interior rooms. All images are resized to 224$\times$224.

\noindent\textcolor{cl4}{$\bullet$} \textbf{KITTI Opt. Flow}~\cite{Geiger2012CVPR} - This is an optical flow prediction task from two consecutive frames. The data comes from a self driving dataset. Crops of size 513$\times$513 sampled from the full image are used during training, and evaluation is done at full resolution.

The following tasks have been adopted from VTAB \cite{vtab}: Caltech Cls., CIFAR-100 Cls., dtd Cls., Pets Cls., SUN Scene Cls., EuroSAT Cls., and CLEVR Count.

\section{End Task Networks}
\label{app:networks}
\begin{figure*}[h]
    \centering
    \includegraphics[width=42pc]{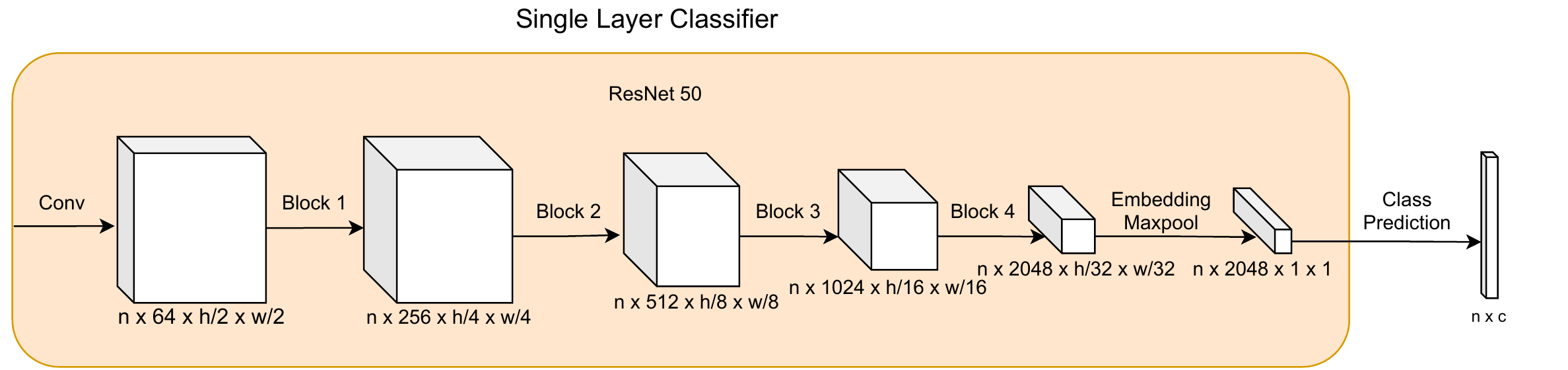}
    \caption{The Single Layer Classifier end task architecture. The orange box shows the frozen encoder.}
    \label{fig:single_layer}
\end{figure*}

\begin{figure*}[h]
    \centering
    \includegraphics[width=42pc]{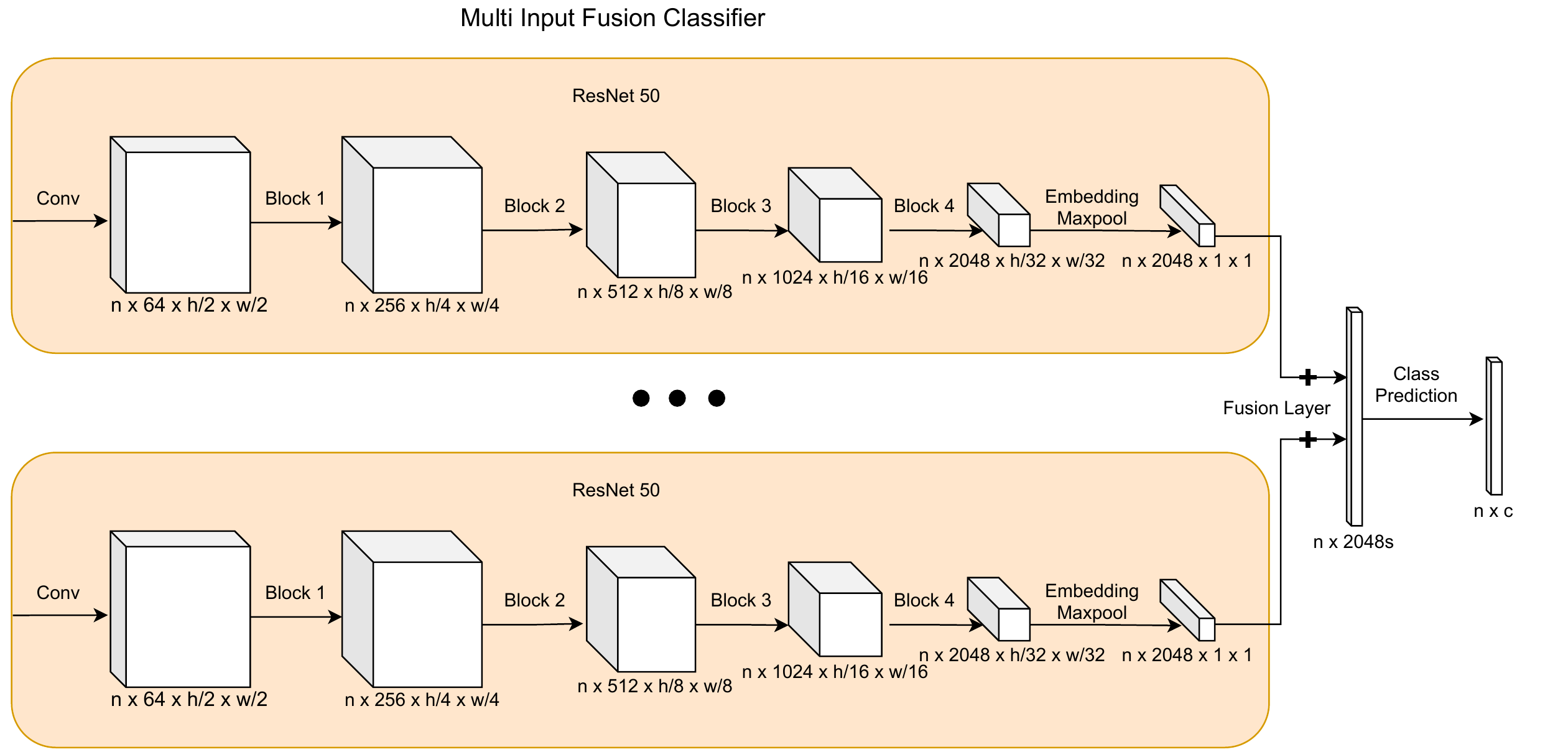}
    \caption{The Multi Input Fusion Classifier end task architecture. The orange box shows the frozen encoder.}
    \label{fig:multi-input}
\end{figure*}

\begin{figure*}[h]
    \centering
    \includegraphics[width=42pc]{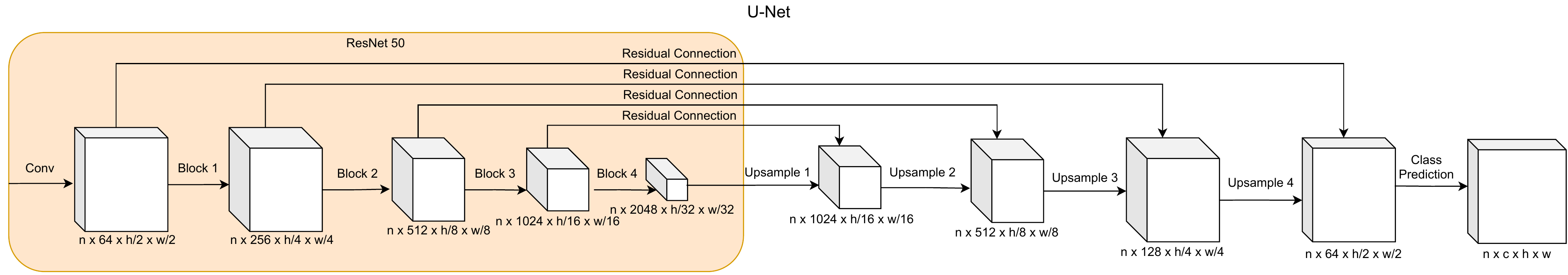}
    \caption{The U-Net end task architecture. The orange box shows the frozen encoder.}
    \label{fig:unet}
\end{figure*}

\begin{figure*}[h]
    \centering
    \includegraphics[width=42pc]{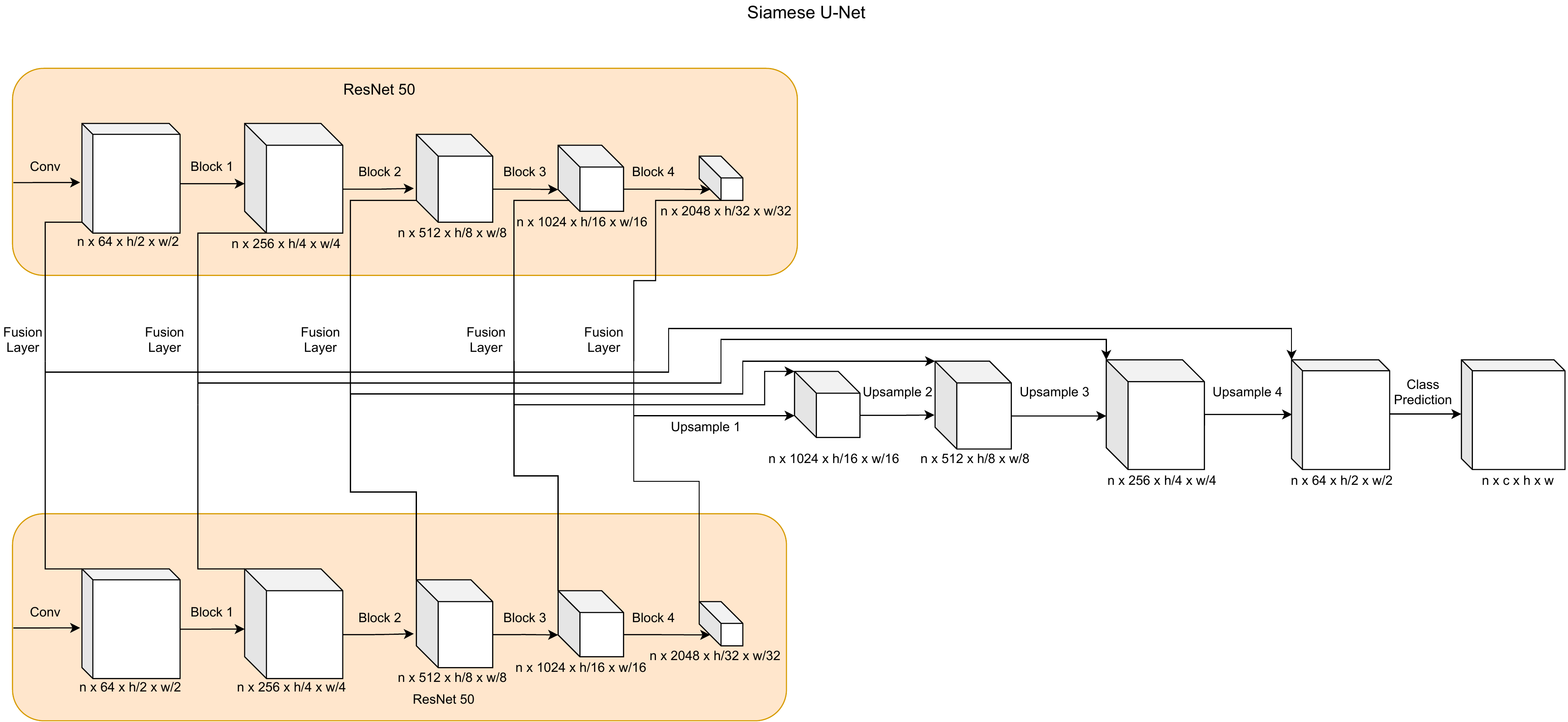}
    \caption{The Siamese U-Net end task architecture. The orange box shows the frozen encoder.}
    \label{fig:siamese-unet}
\end{figure*}

\begin{figure*}[h]
    \centering
    \includegraphics[width=42pc]{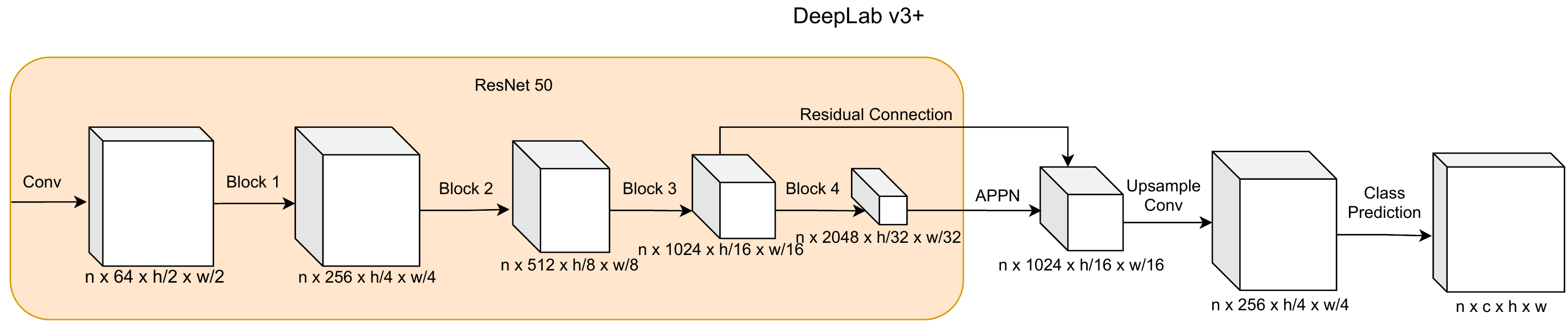}
    \caption{The DeepLabV3+ end task architecture. The orange box shows the frozen encoder.}
    \label{fig:deeplab}
\end{figure*}

The architecture and the loss functions used for each end task have been shown in Table~\ref{tab:task-details}. Top-1 accuracy is the percentage of test samples labeled with the correct class, mIOU is the class wise average intersection over union between the prediction class mask and the ground truth, Negative L1 Error is the negative absolute distance between the prediction and the label averaged over all the pixels, and 1-All is 1 minus the percentage of outliers averaged over all ground truth pixels. \Cref{fig:single_layer,fig:multi-input,fig:unet,fig:siamese-unet,fig:deeplab} show the details of each network. The orange box in the figures shows the frozen encoder. The output dimensions for each block are also shown. The variables $h$ and $w$ represent the height and width of the input image, respectively, while $n$ represents the batch size and $s$ represents the sequence length.

\section{Training Details}
\label{app:train_details}
In this work encoders and end task networks are trained separately. Below we describe the training procedure for each.

We train the encoders by MoCov2 \cite{mocov2} and SwAV \cite{swav} algorithms. For the rest of the training algorithms, we use the publicly released weights for the trained models. We train every model using code publicly released by the authors and the same hyperparameters as the original implementation.

We train the end task networks by freezing the encoders and training just the end task network layers. For each task, we perform a grid search of 4 sets of optimizers and learning rates using the encoder trained with SwAV on ImageNet for 200 epochs. We then select the best performing set of hyperparameters and use them for all other runs. We also use the grid search training runs to determine the number of epochs necessary for each task to converge. We performed grid search for each individual encoder on a subset of all the tasks and found that the hyperparameters we found were the same across all encoders for almost all tasks (and where they were not the same, the performance difference was so small it could be attributed to noise), so due to computation constrains we decided to not perform a full grid search for every task and every model. In Table~\ref{tab:train-details} we report the specific hyperparameters used for each end task.

\section{Correlation Analysis of the End Tasks}
\label{app:correlation}
To better understand the relationships between the end tasks chosen for this paper, we analyze the correlation between their performances using different encoders. Specifically, for every task $A$ and every task $B$ we compute the correlation between the performance of task $A$ and $B$ of all of the encoders we analyze. This shows whether good performance on one task is indicative of good performance on another. 

\begin{figure*}[htb!]
    \centering
    \includegraphics[width=40pc]{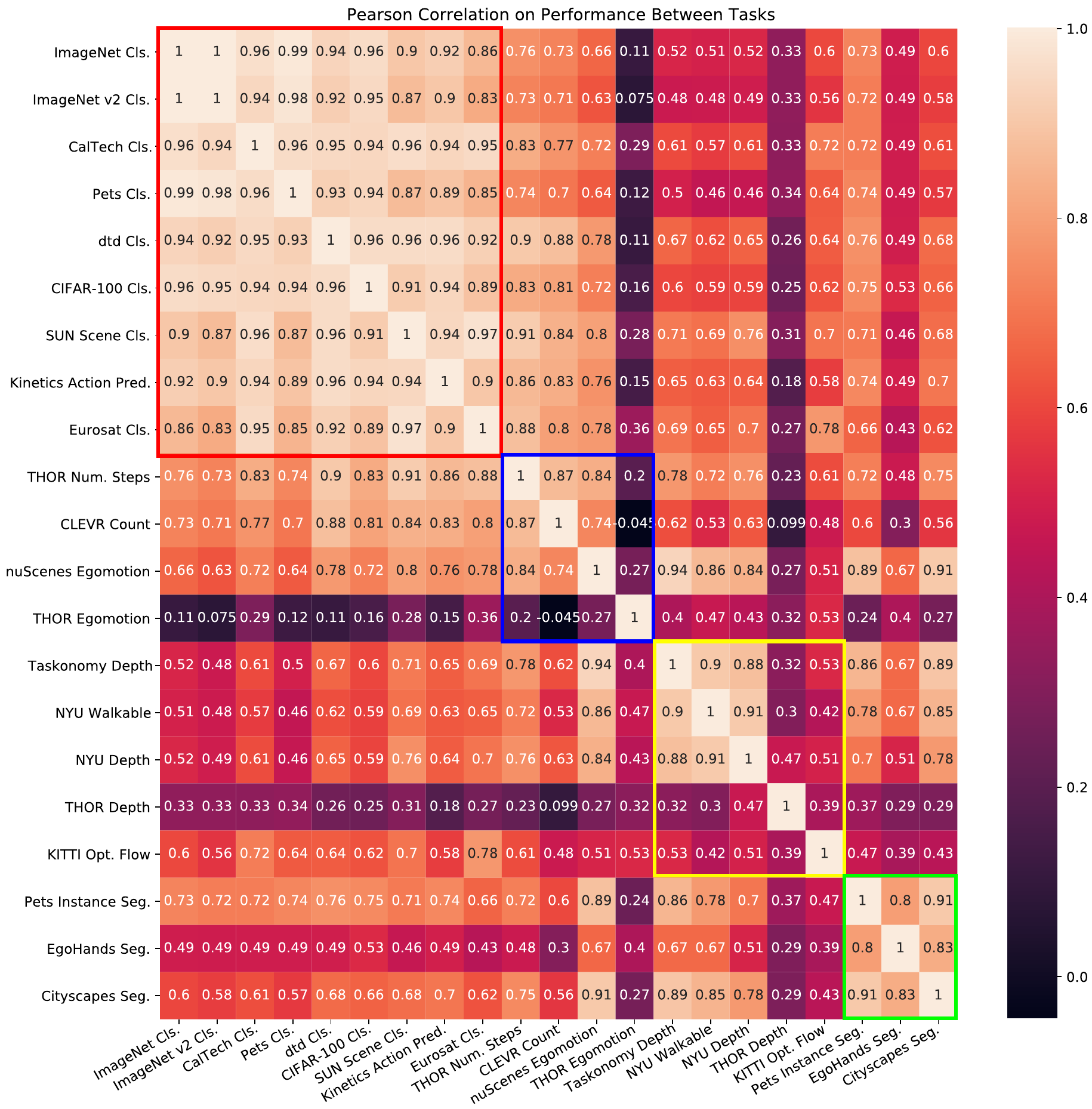}
    \caption{\textbf{Pearson Correlation} between the scores of all end tasks obtained using every encoder we study in this paper. Within category correlations for Semantic Image-level, Structural Image-level, Structural Pixelwise, and Semantic Pixelwise tasks are highlighted by red, blue, yellow and green boxes, respectively.}
    \label{fig:pearson}
\end{figure*}

\begin{figure*}[htb!]
    \centering
    \includegraphics[width=40pc]{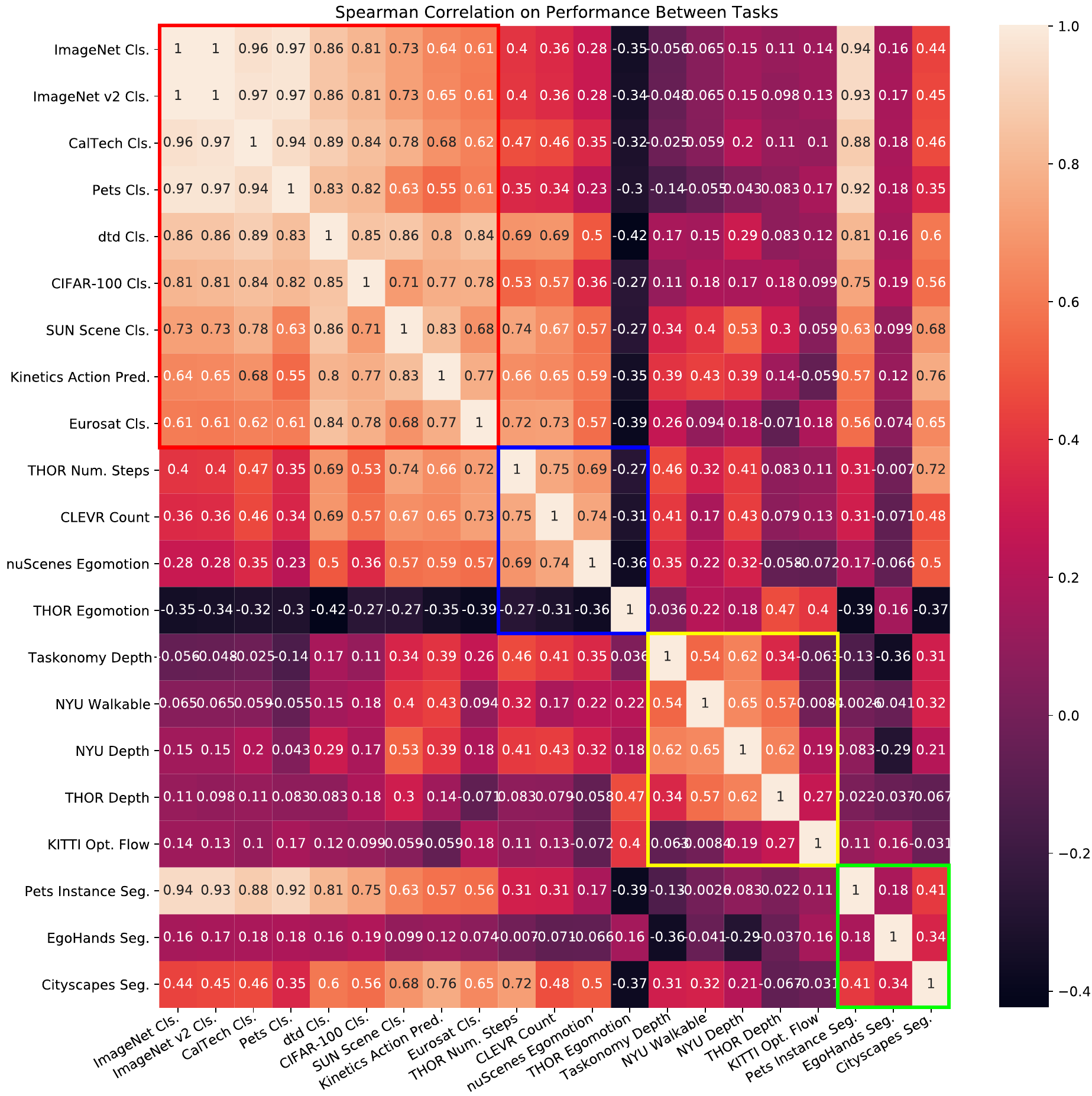}
    \caption{\textbf{Spearman Correlation} between the scores of all end tasks obtained using every encoder we study in this paper. Within category correlations for Semantic Image-level, Structural Image-level, Structural Pixelwise, and Semantic Pixelwise tasks are highlighted by red, blue, yellow and green boxes, respectively.}
    \label{fig:spearman}
\end{figure*}

Figures~\ref{fig:pearson} and \ref{fig:spearman} show the Pearson and Spearman (rank) correlations between the end task performance of the encoders. One clear trend is that we see pockets of strong correlation within each task category. Sometimes they are well defined (Semantic Image-level or Structural Pixelwise tasks represented by red and yellow boxes in Figure~\ref{fig:spearman}) and sometimes they are more subtle (Semantic Pixelwise represented by the green box in Figure~\ref{fig:spearman}). Another trend that these figures show is that ImageNet classification performance is not a good universal metric for encoder performance (especially for pixelwise output tasks, where there is a low correlation). 

\section{CKA Analysis Details}
\label{app:cka}
Centered Kernel Alignment \cite{CKA} is a method of quantifying the similarity of representations between images as they are processed through an encoder. For this study we compare how the relationship between the representations of two images change across the different blocks of the ResNet encoder. We select a balanced subset of 10,000 images from the ImageNet dataset to measure the similarity of representations, and downscale the images to 112$\times$112 before processing them through the encoder. We then compute the CKA between the representations of every pair of images in our subset for every block of the ResNet encoder (this similarity metric has a range of 0 to 1). We find that all encoders trained with the MoCov2 algorithm have an average increase of 0.18 of the average correlation between the layers versus the encoders trained with the SwAV algorithm. This indicates that the MoCov2 encoders retain more spatial information about the images in the later layers and offers a potential hypothesis as to why MoCov2 encoders tend to outperform SwAV encoders at pixelwise output tasks.

It is important to note that this analysis was performed using only a subsample of ImageNet data. ImageNet was chosen for this analysis as it is amongst the most diverse datasets utilized in this paper, but it makes this analysis far from entirely comprehensive. The reason for running this analysis on just this subsample was computational complexity, as evaluating the CKA on all the data available to us is computationally impractical.

\section{ANOVA Tests}
\label{app:anova}
For this test, we consider encoders trained with the MoCov2 and SwAV algorithms on subsets of ImageNet (as discussed in the main text). We examine the relationship between encoders trained on class unbalanced versions of ImageNet and their balanced counterparts with an equivalent number of samples. We use the end task results of the following encoders in our analysis: SwAV Half ImageNet 200, SwAV Linear Unbalanced ImageNet 200, SwAV Quarter ImageNet 200, SwAV Log Unbalanced ImageNet 200, MoCov2 Half ImageNet 200, MoCov2 Linear Unbalanced ImageNet 200, MoCov2 Quarter ImageNet 200, MoCov2 Log Unbalanced ImageNet 200.

Our analysis found evidence that an encoder trained on a Log Unbalanced subset of ImageNet outperforms an encoder trained on a balanced subset of ImageNet with an equivalent number of samples. To further validate this conclusion we trained 2 additional encoders using SwAV on 2 different logarithmically unbalanced subsets of ImageNet and included them in the following test.

We fit an ANOVA model to all of the results we obtain, treating the task, training algorithm, dataset balance, dataset size and number of training steps as variables. We find that (unsurprisingly) the task, dataset size and number of training steps are statistically significant indicators of end task performance. We also find that the algorithm used to train the encoder (MoCov2 vs SwAV) is a statistically significant indicator of end task performance, with SwAV models performing better (this does not contradict our claim that SwAV is not universally better than MoCov2, as we simply have more tasks that SwAV is good at in our test battery). Finally, we do not find any statistically significant evidence that an encoder trained with the balanced ImageNet is better than the encoders trained on the discussed unbalanced variations. We do however find evidence that an encoder trained on a Log unbalanced subset of ImageNet tends to perform better than one trained on a balanced subset. Perhaps the (comparatively) larger number of samples of the same few categories is a good match for the contrastive learning algorithm, but further experiments are needed to determine the exact cause and extent of this phenomenon.

\section{Variance of the Results}

The main source of variance in our results is the self-supervised training of the encoder. Since each encoder requires over 500 GPU hours to be trained for 200 epochs with the MoCov2 training algorithm, and over 1000 GPU hours to be trained for 200 epochs with the SwAV training algorithm, it is impractical for us to test multiple training runs of every encoder configuration that we study in this work.

To provide some context regarding the magnitude of variations across runs, we train three encoders using SwAV on ImageNet for 200 epoch with different random seeds. All training parameters are exactly the same as those used by the SwAV authors to obtain their SwAV 200 model. 

Our results show that, on average, the variation in the performance of the end tasks is less than 0.85\% (relative difference with the average performance), which can be negligible. 

\section{List of Encoders}
\label{app:encoders}
Table~\ref{tab:encoder-list} provides a complete list of all 30 encoders that are used for our analysis. 

\section{Effects of MultiCrop Pre-processing}
\label{sec:multicrop}
This work draws some comparisons between the MoCov2 and SwAV training pipelines and identifies some trends in the performance of encoders trained with them. 

The two pipelines do not just contain a different training algorithm, but they also employ different pre-processing methods. To understand if the observed differences in end task performance are simply a result of different pre-processing we conduct an ablation study where we use the improved pre-processing methods of SwAV in conjunction with the MoCov2 training pipeline to train an encoder on ImageNet and evaluate its performance on our battery of end tasks. 

We observe that the MultiCrop pre-procesing employed by SwAV is only partially responsible for the observed gap between the two training pipelines in question. Furthermore we observe that the MuliCrop pre-processing is not a universally better choice, as it seems to degrade the performance of certain Pixelwise output tasks. This result is rather expected since the MultiCrop pre-processing essentially makes the model embed a patch of the image and the entire image very similarly, thus encouraging more semantic and less structural embeddings. 

\begin{figure*}[h]
    \centering
    \includegraphics[width=42pc]{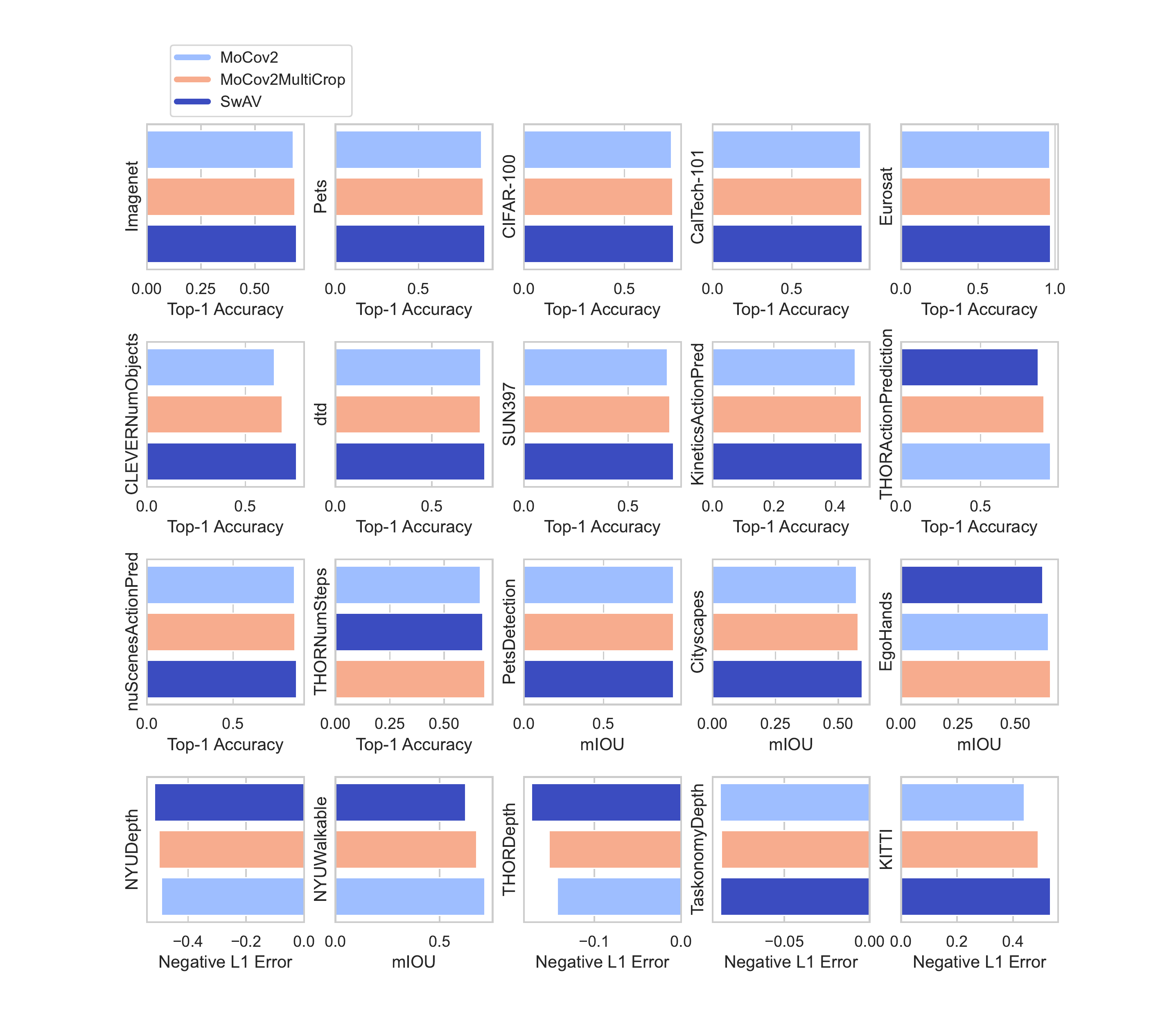}
    \caption{End Task performance of encoders trained on ImageNet using MoCov2, SwAV and MoCov2 with the MultiCrop pre-processing step from the SwAV paper. Different end tasks have different measures of performance but a higher number always indicates better performance. The bars represent the performance of different encoders and are sorted from least to most performant on each end task.}
    \label{fig:multi_crop}
\end{figure*}

Figure ~\ref{fig:multi_crop} shows that for almost all tasks the performance of the MoCov2+MultiCrop model is between that of the SwAV model and the vanilla MoCov2. From this we can hypothesize that adding MultiCrop makes the MoCov2 model behave more like model trained with SwAV when embedding images.

\section{Other Encoders} 
\label{sec:addencoder}
One obvious axis of expansion for future work is performing this analysis on more encoders trained with different pipelines. We chose a very small subset from the current state of the field and analyzed them very comprehensively. This meant that we would necessarily have to omit some prominent pipelines from our study. We conducted small ablations with 2 such noteworthy omissions: SimSiam~\cite{chen2020simsiam}, a siamese-style self supervised algorithm and Exemplar-v2~\cite{zhao2021what}, an improved supervised training method. 

\begin{figure*}[h]
    \centering
    \includegraphics[width=42pc]{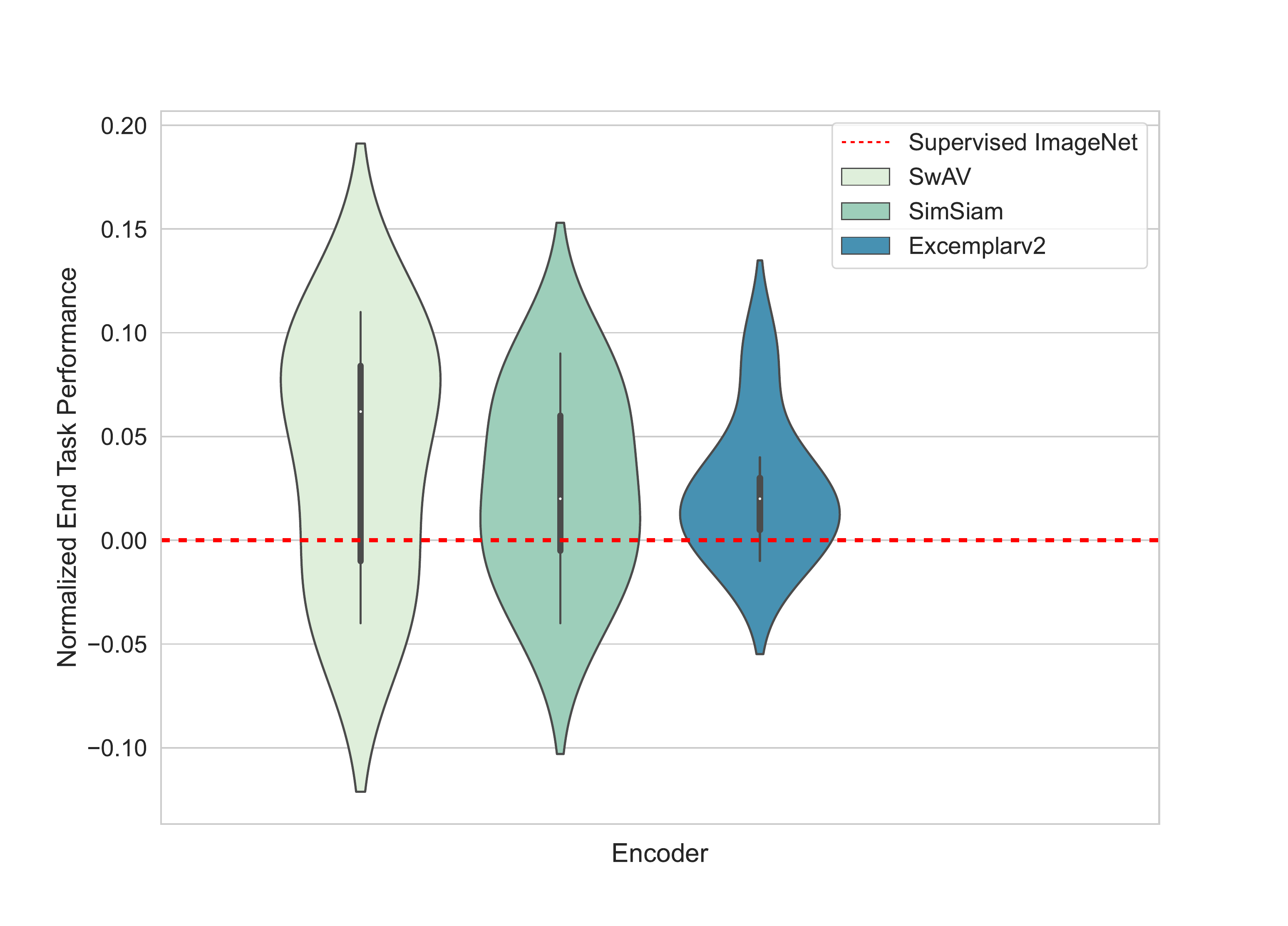}
    \caption{Distribution of normalized performances for the SwAV, SimSiam and Exemplar-v2 encoders trained on ImageNet for 200 epochs. The performances are normalized by first subtracting the performance of the vanilla supervised ImageNet encoder and then dividing by the standard deviation of all the performances for the task.  Positive values show superior performance to the vanilla supervised ImageNet encoder, and the negative values show otherwise. A larger width means more performance values fall in that range. The plot shows that SimSiam tends to perform similarly to SwAV. Furthermore the plot shows that Exemplar-v2 performs better than the vanilla baseline, but worse than both SwAV and SimSiam, reinforcing our claims about the outperformance of self-supervised models.}
    \label{fig:excemplar_simsiam}
\end{figure*}

Figure~\ref{fig:excemplar_simsiam} shows that SimSiam performs very similarly to SwAV on our battery of end tasks. The distributions of the normalized end task scores of SwAV and SimSiam encoders show that SimSiam does not appear to be better and thus our analysis did not miss covering an encoder that would significantly outperform the rest.

We can also see that Exemplar-v2 does in fact outperform the vanilla supervised baseline on most end tasks, but it falls far short of the performance of certain self-supervised models like SwAV. This suggests that our findings regarding the performance of supervised vs. self supervised pipelines still hold.

\begin{table*}[tp]\small
\centering
\begin{tabular}{|l|c|c|c|c|}
\hline
\textbf{Task} & \textbf{Category} & \textbf{End Task Network} & \textbf{Loss} & \textbf{Success Metric} \\ 
\hline\hline
\textcolor{cl1}{$\bullet$} ImageNet Cls. & Semantic Image-level & Single Layer Classifier & Cross Entropy & Top-1 Accuracy\\
\hline
\textcolor{cl1}{$\bullet$} ImageNet v2 Cls. &  Semantic Image-level & Single Layer Classifier & Cross Entropy & Top-1 Accuracy \\
\hline
\textcolor{cl1}{$\bullet$} Pets Cls. &  Semantic Image-level & Single Layer Classifier & Cross Entropy & Top-1 Accuracy \\
\hline
\textcolor{cl1}{$\bullet$} CalTech Cls. &  Semantic Image-level & Single Layer Classifier & Cross Entropy & Top-1 Accuracy \\
\hline
\textcolor{cl1}{$\bullet$} CIFAR-100 Cls. &  Semantic Image-level & Single Layer Classifier & Cross Entropy & Top-1 Accuracy \\
\hline
\textcolor{cl1}{$\bullet$} SUN Scene Cls. &  Semantic Image-level & Single Layer Classifier & Cross Entropy & Top-1 Accuracy \\
\hline
\textcolor{cl1}{$\bullet$} Eurosat Cls. &  Semantic Image-level & Single Layer Classifier & Cross Entropy & Top-1 Accuracy \\
\hline
\textcolor{cl1}{$\bullet$} dtd Cls. &  Semantic Image-level & Single Layer Classifier & Cross Entropy & Top-1 Accuracy \\
\hline
\textcolor{cl1}{$\bullet$} Kinetics Action Pred. &  Semantic Image-level &  Multi Input Fusion Classifier & Cross Entropy & Top-1 Accuracy \\
\hline
\textcolor{cl2}{$\bullet$} CLEVR Count &  Structural Image-level & Single Layer Classifier & Cross Entropy & Top-1 Accuracy \\
\hline
\textcolor{cl2}{$\bullet$} THOR Num. Steps & Structural Image-level & Single Layer Classifier & Cross Entropy & Top-1 Accuracy \\
\hline
\textcolor{cl2}{$\bullet$} THOR Egomotion & Structural Image-level & Multi Input Fusion Classifier & Cross Entropy & Top-1 Accuracy \\
\hline
\textcolor{cl2}{$\bullet$} nuScenes Egomotion & Structural Image-level & Multi Input Fusion Classifier & Cross Entropy & Top-1 Accuracy \\
\hline
\textcolor{cl3}{$\bullet$} Cityscapes Seg. & Semantic Pixelwise & DeepLabv3+ & Pixelwise Cross Entropy & mIOU \\
\hline
\textcolor{cl3}{$\bullet$} Pets Instance Seg. & Semantic Pixelwise &  DeepLabv3+ & Pixelwise Cross Entropy & mIOU \\
\hline
\textcolor{cl3}{$\bullet$} EgoHands Seg. & Semantic Pixelwise &  DeepLabv3+ & Pixelwise Cross Entropy & mIOU \\
\hline
\textcolor{cl4}{$\bullet$} THOR Depth & Structural Pixelwise & U-Net & L1 Error & Negative L1 Error \\
\hline
\textcolor{cl4}{$\bullet$} Taskonomy Depth & Structural Pixelwise & U-Net & L1 Error & Negative L1 Error \\
\hline
\textcolor{cl4}{$\bullet$} NYU Depth & Structural Pixelwise & U-Net & L1 Error & Negative L1 Error \\
\hline
\textcolor{cl4}{$\bullet$} NYU Walkable & Structural Pixelwise &  DeepLabv3+ & Pixelwise Cross Entropy & mIOU \\
\hline
\textcolor{cl4}{$\bullet$} KITTI Opt. Flow & Structural Pixelwise & Siamese U-Net. & L1 Error & 1 - All \\
\hline
\end{tabular}
\caption{The network architecture, the loss and the success metric for each end task.}
\label{tab:task-details}
\end{table*}


\begin{table*}[tp]\small
\centering
\begin{tabular}{|l|r|c|c|c|}
\hline
\textbf{Task} & \textbf{Train Set Size} & \textbf{Number of Train Epochs} & \textbf{Optimizer} & \textbf{Learning Rate} \\ 
\hline\hline
\textcolor{cl1}{$\bullet$} ImageNet Cls. & 1,281,167 & 100 & Adam & 0.0003\\
\hline
\textcolor{cl1}{$\bullet$} Pets Cls. & 3,680  & 500 & Adam & 0.0003 \\
\hline
\textcolor{cl1}{$\bullet$} CalTech Cls. &  3,060  & 5000 & Adam& 0.0003 \\
\hline
\textcolor{cl1}{$\bullet$} CIFAR-100 Cls. &  50,000 & 100 & Adam& 0.0003 \\
\hline
\textcolor{cl1}{$\bullet$} SUN Scene Cls. &  87,003 & 250 & Adam & 0.0003 \\
\hline
\textcolor{cl1}{$\bullet$} Eurosat Cls. &  21,600 & 200 & Adam & 0.0003 \\
\hline
\textcolor{cl1}{$\bullet$} dtd Cls. &  3,760 & 100 & Adam & 0.0003 \\
\hline
\textcolor{cl1}{$\bullet$} Kinetics Action Pred. &  50,000 & 100 & Adam & 0.0003 \\
\hline
\textcolor{cl2}{$\bullet$} CLEVR Count & 70,000 & 100 & Adam & 0.0003 \\
\hline
\textcolor{cl2}{$\bullet$} THOR Num. Steps & 60,000 & 100 & Adam & 0.0003 \\
\hline
\textcolor{cl2}{$\bullet$} THOR Egomotion & 60,000 & 100 & Adam & 0.0003 \\
\hline
\textcolor{cl2}{$\bullet$} nuScenes Egomotion & 28,000 & 100 & Adam & 0.0003 \\
\hline
\textcolor{cl3}{$\bullet$} Cityscapes Seg. & 3,475 & 100 & Adam & 0.0003 \\
\hline
\textcolor{cl3}{$\bullet$} Pets Instance Seg. & 3,680 & 100 & Adam & 0.0003 \\
\hline
\textcolor{cl3}{$\bullet$} EgoHands Seg. & 4,800 &  25 & Adam & 0.0003 \\
\hline
\textcolor{cl4}{$\bullet$} THOR Depth & 60,000 & 50 & Adam & 0.0003 \\
\hline
\textcolor{cl4}{$\bullet$} Taskonomy Depth & 39,995 & 50 & Adam & 0.0003 \\
\hline
\textcolor{cl4}{$\bullet$} NYU Depth & 1,159 & 250 & Adam & 0.0003 \\
\hline
\textcolor{cl4}{$\bullet$} NYU Walkable & 1,159  & 100 & Adam & 0.0003 \\
\hline
\textcolor{cl4}{$\bullet$} KITTI Opt. Flow & 200 & 250 & Adam & 0.0003 \\
\hline
\end{tabular}
\caption{Training details for each end task.}
\label{tab:train-details}
\end{table*}


\begin{table*}[tp]\small
\centering
\begin{tabular}{|l|r|c|c|c|c|}
\hline
\textbf{Encoder Name} & \textbf{Method} & \textbf{Dataset} & \textbf{Dataset Size} & \textbf{Number of Epochs} & \textbf{Trained by us} \\ 
\hline\hline
SwAV ImageNet 800 & SwAV & ImageNet & 1.3M & 800 & No\\
\hline
SwAV ImageNet 200 & SwAV & ImageNet & 1.3M & 200 & No\\
\hline
SwAV ImageNet 100 & SwAV & ImageNet & 1.3M & 100 & Yes\\
\hline
SwAV ImageNet 50 & SwAV & ImageNet & 1.3M & 50 & Yes\\
\hline
SwAV Half ImageNet 200 & SwAV & ImageNet-\sfrac{1}{2} & 0.5M & 200 & Yes\\
\hline
SwAV Half ImageNet 100 & SwAV & ImageNet-\sfrac{1}{2} & 0.5M & 100 & Yes\\
\hline
SwAV Quarter ImageNet 200 & SwAV & ImageNet-\sfrac{1}{4} & 0.25M & 200 & Yes\\
\hline
SwAV Linear Unbalanced ImageNet 200 & SwAV & ImageNet-\sfrac{1}{2}-Lin & 0.5M & 200 & Yes\\
\hline
SwAV Linear Unbalanced ImageNet 100 & SwAV & ImageNet-\sfrac{1}{2}-Lin & 0.5M & 100 & Yes\\
\hline
SwAV Log Unbalanced ImageNet 200 & SwAV & ImageNet-\sfrac{1}{4}-Log & 0.25M & 200 & Yes\\
\hline
SwAV Places 200 & SwAV & Places & 1.3M & 200 & Yes\\
\hline
SwAV Kinetics 200 & SwAV & Kinetics & 1.3M & 200 & Yes\\
\hline
SwAV Taskonomy 200 & SwAV & Taskonomy & 1.3M & 200 & Yes\\
\hline
SwAV Combination 200 & SwAV & Combination & 1.3M & 200 & Yes\\
\hline
MoCov2 ImageNet 800 & MoCov2 & ImageNet & 1.3M & 800 & No\\
\hline
MoCov2 ImageNet 200 & MoCov2 & ImageNet & 1.3M & 200 & No\\
\hline
MoCov2 ImageNet 100 & MoCov2 & ImageNet & 1.3M & 100 & Yes\\
\hline
MoCov2 ImageNet 50 & MoCov2 & ImageNet & 1.3M & 50 & Yes\\
\hline
MoCov2 Half ImageNet 200 & MoCov2 & ImageNet-\sfrac{1}{2} & 0.5M & 200 & Yes\\
\hline
MoCov2 Half ImageNet 100 & MoCov2 & ImageNet-\sfrac{1}{2} & 0.5M & 100 & Yes\\
\hline
MoCov2 Quarter ImageNet 200 & MoCov2 & ImageNet-\sfrac{1}{4} & 0.25M & 200 & Yes\\
\hline
MoCov2 Linear Unbalanced ImageNet 200 & MoCov2 & ImageNet-\sfrac{1}{2}-Lin & 0.5M & 200 & Yes\\
\hline
MoCov2 Linear Unbalanced ImageNet 100 & MoCov2 & ImageNet-\sfrac{1}{2}-Lin & 0.5M & 100 & Yes\\
\hline
MoCov2 Log Unbalanced ImageNet 200 & MoCov2 & ImageNet-\sfrac{1}{4}-Log & 0.25M & 200 & Yes\\
\hline
MoCov2 Places 200 & MoCov2 & Places & 1.3M & 200 & Yes\\
\hline
MoCov2 Kinetics 200 & MoCov2 & Kinetics & 1.3M & 200 & Yes\\
\hline
MoCov2 Taskonomy 200 & MoCov2 & Taskonomy & 1.3M & 200 & Yes\\
\hline
MoCov2 Combination 200 & MoCov2 & Combination & 1.3M & 200 & Yes\\
\hline
MoCov1 ImageNet 200 & MoCov1 & ImageNet & 1.3M & 200 & No\\
\hline
PIRL ImageNet 800 & PIRL & ImageNet & 1.3M & 800 & No\\
\hline
\end{tabular}
\caption{The complete list of all 30 encoders used for the study.}
\label{tab:encoder-list}
\end{table*}

\FloatBarrier
\clearpage

\end{document}